\documentclass[10pt,letterpaper]{article}
\usepackage[top=0.85in,left=2.75in,footskip=0.75in]{geometry}

\usepackage{amsmath,amssymb}

\usepackage{changepage}

\usepackage[utf8x]{inputenc}

\usepackage{textcomp,marvosym}

\usepackage{cite}

\usepackage{nameref,hyperref}

\usepackage[right]{lineno}

\newcommand{\emoji}[1]{\includegraphics[width=1em]{emoji_images/#1.png}}
\usepackage{booktabs}

\usepackage{microtype}
\DisableLigatures[f]{encoding = *, family = * }

\usepackage[table]{xcolor}

\usepackage{array}
\usepackage{soul}

\newcolumntype{+}{!{\vrule width 2pt}}

\newlength\savedwidth

\raggedright
\usepackage[aboveskip=1pt,labelfont=bf,labelsep=period,justification=raggedright,singlelinecheck=off]{caption}

\makeatletter
\renewcommand{\@biblabel}[1]{\quad#1.}
\makeatother

\usepackage{comment}

\usepackage{lastpage,fancyhdr,graphicx}
\usepackage{epstopdf}
\fancyheadoffset[L]{2.25in}
\fancyfootoffset[L]{2.25in}
\begin{document}
\vspace*{0.2in}

\begin{flushleft}
{\Large
\textbf\newline{Emojis predict dropouts of remote workers: An empirical study of emoji usage on GitHub} 
}
\newline
\\
Xuan Lu\textsuperscript{1},
Wei Ai\textsuperscript{2},
Zhenpeng Chen\textsuperscript{3},
Yanbin Cao\textsuperscript{3},
Qiaozhu Mei\textsuperscript{1*}
\\
\bigskip
\textbf{1} School of Information, University of Michigan, Ann Arbor, Michigan, USA
\\
\textbf{2} College of Information Studies, University of Maryland, College Park, Maryland, USA
\\
\textbf{3} Peking University, Beijing, China
\\
\bigskip

%
%





* qmei@umich.edu

\end{flushleft}
\section*{Abstract}
Emotions at work have long been identified as critical signals of work motivations, status, and attitudes, and as predictors of various work-related outcomes. When more and more employees work remotely, these emotional signals of workers become harder to observe through daily, face-to-face communications. 

The use of online platforms to communicate and collaborate at work provides an alternative channel to monitor the emotions of workers. This paper studies how emojis, as non-verbal cues in online communications, can be used for such purposes and how the emotional signals in emoji usage can be used to predict future behavior of workers. In particular, we present how the developers on GitHub use emojis in their work-related activities.  We show that developers have diverse patterns of emoji usage, which can be related to their working status including activity levels, types of work, types of communications, time management, and other behavioral patterns.  Developers who use emojis in their posts are significantly less likely to dropout from the online work platform. Surprisingly, solely using emoji usage as features, standard machine learning models can predict future dropouts of developers at a satisfactory accuracy. Features related to the general use and the emotions of emojis appear to be important factors, while they do not rule out paths through other purposes of emoji use.


\section*{Introduction}

The future of work is less dependent on centralized workplaces and in-person collaborations. Indeed, working remotely has become the new norm through the COVID-19 pandemic, and the trend is likely to continue afterward. The IT industry, naturally more familiar with virtual platforms and having fewer logistic constraints, has been leading the tide by even allowing employees to work from home permanently~\cite{bbc2020microsoft}.

While to what extent working remotely influences work outcomes is still debatable, a more visible concern is its impact on working (and organizational) practices. A good example is to handle emotions at work, which have long been identified as critical signals of motivations, attitudes, and mental health status of workers, as well as culture, organizational justices, and other environmental factors of workplaces (e.g.,~\cite{Rafaeli1989-ma, Grandey2008-hz, Farh2012-pl}). Workers' emotions, especially those indicating their mental health status (e.g., ~\cite{Staw1994-lx, Thoresen2003-wt, carayon1999work}), have been linked to various short-term and long-term work-related outcomes, such as engagement through paths of hope~\cite{Ouweneel2012-hx}, goal commitments, and motivation~\cite{Seo2004-zv}. Voluntary work may be more ``affect-driven'' than in-role work~\cite{Grandey2008-hz}. 
Harmonious passion increases engagement~\cite{vallerand2003passions}, satisfaction, and creativity~\cite{liu2011autonomy}; obsessive passion leads to burnouts~\cite{lavigne2012passion}; and burnouts lead to lower productivity, conflicts, and dropouts from work~\cite{schaufeli1998burnout}. Stress at work may even cause physical and mental disorders in life~\cite{carayon1999work}. In non-work related contexts such as social communities \cite{wang2012stay} and online education \cite{dillon2016student}, emotions and emotional support have also been identified as critical factors for the retention of members and students. 

At in-person workplaces, such signals can be frequently observed from the emotions that workers express during face-to-face communications. These signals are much harder to observe at virtual workplaces.  A recent report by GitHub~\cite{github2020spotlighht} notices that developers work for longer hours during the pandemic, which raises  concerns about potential burnouts. This implication, however, could not be verified or falsified without a measurement of the developers' true emotional status. How to track emotions of remote workers is a major challenge for the future of work.   

Our work is motivated to address this challenge. We notice that remote workers rely heavily on online platforms to collaborate and to communicate with each other. While some of the communications are live and face-to-face (e.g., video conferences), more are asynchronous and textual (e.g., emails, instant messages, forum discussions, Slack chats). If workers frequently express emotions in these contexts in a similar way to face-to-face communications, then one can potentially monitor these channels to track their emotional status, possibly with the help of a data mining system. 

Questions are whether workers do frequently express emotions in online work-related communications, and if they do, whether these emotions are easily trackable, and whether they are representative of work-related status (rather than personality and  preferences). More importantly, are such online emotional signals predictive of any work-related outcomes, similar to how passion, stress, and burnouts are able to predict work engagement and attrition?   

Answering these questions requires analyzing longitudinal datasets of activities and communications on online work platforms.  The analysis also relies on identifying clean emotional signals in online communications.  While one could use natural language processing techniques to extract emotions and sometimes infer mental health issues from text~\cite{chen2018mood, calvo2017natural, de2014mental}, the ambiguity of free text often makes it hard to distinguish emotions from work-related content.  Text also fails to capture non-verbal signals (e.g., facial expressions and gestures), which are so commonly used to express emotions in-person. 

The key innovation of our study is to use emojis as sensors of emotions of remote workers. Emojis are widely studied as non-verbal cues in text \cite{Miller2017-il}, carriers of emotions \cite{donovan2016mood}, a new ubiquitous language \cite{Mei2019-qk}, as well as 
sensors of cultural backgrounds~\cite{lu2016learning}, 
personality~\cite{marengo2017assessing}, engagement~\cite{ge2019emoji}, and mental health status~\cite{settanni2015sharing}.  Emojis represent rich non-verbal cues, which makes them better sensors of affections than words~\cite{donovan2016mood}. In this study, we pay special attentions to this function of expressing emotions, not only because it is a major and original function of emojis but also it is easier to identify and measure. In fact, the rich information embedded in emojis makes them suitable for various purposes of usage beyond expressing emotions, such as to adjust tones~\cite{Cramer2016-ic}, adhere social norms~\cite{Cramer2016-ic}, 
manage conversations~\cite{Kelly2015-ak}, maintain and enhance social relations~\cite{Riordan2017-eo}, and help build one's own identity~\cite{Ge2019-ky}. Emojis may also be used for one or more of these purposes in work-related scenarios and in specific contexts, although they may be harder to distinguish (especially automatically). Therefore, instead of limiting our analysis to only the emojis that express emotions, we also characterize the general usage of emojis and do not intend to rule out the effect of emoji usage for other purposes on work-related outcomes.

We place our analysis on GitHub, the leading online collaborative platform for software development, which has already been used by millions of developers for years. Through a dataset of multi-year event log collected by the GHArchive~\cite{gharchive}, 
we are able to track down multiple types of work activities (such as \textit{push} and \textit{fork}) of developers and their communications (such as \textit{issues} and \textit{commit comments}). We focus our analysis on the usage of emojis in these contexts. 

We are particularly interested in answering the following research questions: (1) how widely do developers use emojis? (2) is the use of emojis related to their working status? (3) can emojis be utilized to predict future dropouts of developers? 

Our analyses reveal that developers frequently use emojis in work-related communications, in a different way from how emojis are used in general social activities. Emoji usage varies among different programming languages. Having programming languages controlled, a comprehensive regression analysis demonstrates that emoji usage is highly related to a developer's working status, including activity levels, types of work, trends of activities, time management, and other behavioral patterns. 

We further investigate the predictive power of emoji usage on an extremely negative outcome: whether an active developer is going to log no activity on GitHub next year (i.e., they drop out from the online work platform). Controlling confounds such as programming languages and activity levels, we find that non-emoji users are \textit{three times more likely} to dropout than emoji users. Using features extracted solely from their emoji usage, standard machine learning models are able to predict the future dropouts of emoji users with impressive accuracy (up to 75\% classification accuracy and 82\% AUC). Features measuring the emotions of emojis indeed appear to be important factors in the prediction, although they do not rule out other potential paths (e.g., other purposes of emoji use). The predictive features are explainable and provide insights on establishing new practices in remote workplaces, such as promoting harmonious and diverse usage of emojis and tracking certain emotional conditions of workers.  

\section*{Description of Emoji Usage on GitHub}\label{sec:descriptive}

We base our study on GitHub, the leading online collaboration platform for software developers.  We collect the event log data released by GHArchive, which includes more than 20 event types provided by GitHub~\cite{webhook}. 
On GitHub, \textit{issues} are used to track ideas, feedback, tasks, or bugs; \textit{pull requests} are used to tell others about changes that have been pushed to a branch in a repository. Collaborators or team members can \textit{comment} on issues and pull requests. Additionally, they can leave comments when they review a pull request or make a commit. 
We denote all \textit{issues}, \textit{pull requests}, and \textit{comments} as \textbf{\textit{posts}}, through which developers on this platform communicate with each other, and all events (including \textit{posts} and operations on them) as work-related \textbf{\textit{activities}} in general. 
In total, we are able to obtain 62,852,221 posts generated in the single year of 2018. 

We approach the first research question, \textbf{\textit{how widely developers use emojis in communications on GitHub}}, in particular, whether developers use emojis frequently in different work activities and across different programming languages. If emojis are widely used at work, it is possible that they may be related to working status of developers and even contain predictive signals for their future behaviors.

\subsection*{Popularity of Emojis on GitHub}\label{subsec:twitter}
About 5.53\% posts on GitHub contain at least one emoji. Is this proportion high? For comparison purposes, we obtain a random sample of 8 billion Tweets from the Twitter Decahose stream in 2018 and find that 17.98\% contain emoji(s). As expected, people are less likely to use emojis in work-related activities than in everyday social activities. However, as shown in Table~\ref{tab:emojis_stats}, the proportion of emoji posts varies between 2.92\% and 14.0\% (which is close to the proportion of emoji in Tweets) in different types of posts on GitHub -- issues, pull requests, their comments, review comments, and commit comments -- which represent different types of work activities. Emojis are noticeably more common among pull request comments and then among commit comments. Considering that both commits and pull requests are related to code submission, their comments are likely to be related to the coordination among developers, and emojis may have been helpful in facilitating such collaborative work. 

\begin{table}[!ht]
    \centering
    \small
    \caption{{\bf Emoji Usage on GitHub by Post Type.} Over 5\% of posts on GitHub contain at least one emoji. Pull request comments and commit comments have higher proportions of emoji posts. }
    \begin{tabular}{crccc|c}
    \toprule
    Type of post & \#Post & \%Emoji post & Top 10 emojis \\\midrule
        Issues & 9.4M & 3.34\% & \emoji{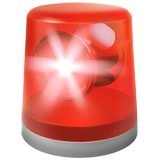} \emoji{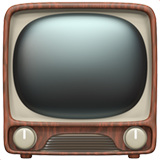} \emoji{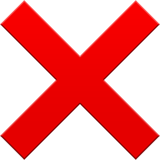} \emoji{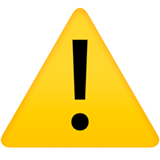} \emoji{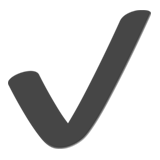} \emoji{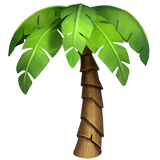} \emoji{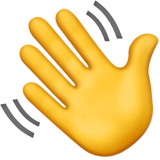} \emoji{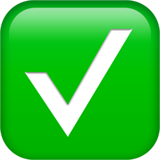} \emoji{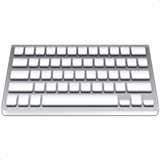} \emoji{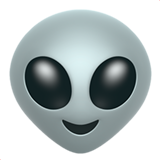}\\
         Issue comments & 18.5M & 3.37\% & \emoji{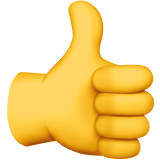} \emoji{keyboard_2328} \emoji{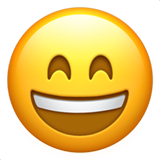} \emoji{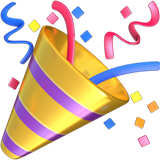} \emoji{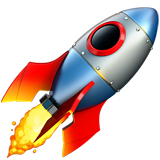} \emoji{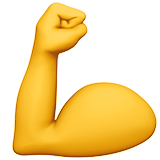} \emoji{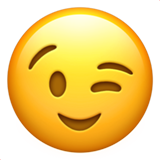} \emoji{police-cars-revolving-light_1f6a8} \emoji{television_1f4fa} \emoji{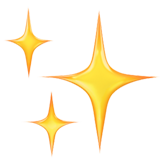}\\
        Pull requests (PR) & 13.9M & 4.14\%  & \emoji{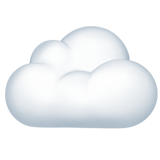} \emoji{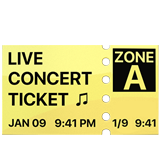} \emoji{rocket_1f680} \emoji{palm-tree_1f334} \emoji{party-popper_1f389} \emoji{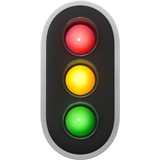} \emoji{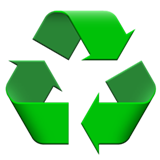} \emoji{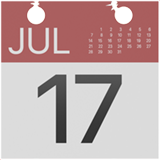} \emoji{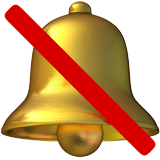} \emoji{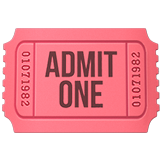} \\ 
        PR comments & 11.9M & 14.0\% & \emoji{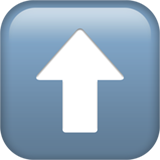} \emoji{rocket_1f680} \emoji{white-heavy-check-mark_2705} \emoji{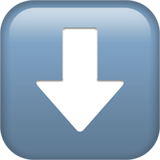} \emoji{thumbs-up-sign_1f44d} \emoji{party-popper_1f389} \emoji{heavy-check-mark_2714} \emoji{cross-mark_274c} \emoji{warning-sign_26a0} \emoji{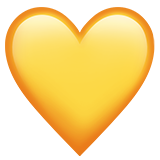}\\
        PR review comments & 8.4M & 2.92\% & \emoji{thumbs-up-sign_1f44d} \emoji{smiling-face-with-open-mouth-and-smiling-eyes_1f604} \emoji{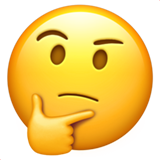} \emoji{winking-face_1f609} \emoji{warning-sign_26a0} \emoji{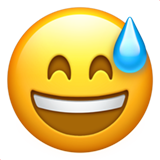} \emoji{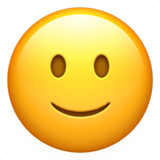} \emoji{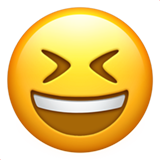} \emoji{white-heavy-check-mark_2705} \emoji{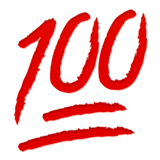}\\
        Commit comments & 0.8M & 6.09\%  & \emoji{white-heavy-check-mark_2705} \emoji{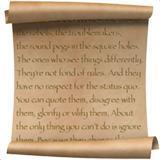} \emoji{thumbs-up-sign_1f44d} \emoji{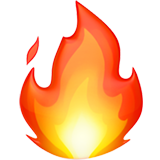} \emoji{hundred-points-symbol_1f4af} \emoji{smiling-face-with-open-mouth-and-smiling-eyes_1f604} \emoji{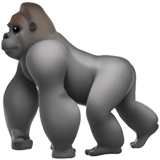} \emoji{party-popper_1f389} \emoji{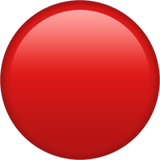} \emoji{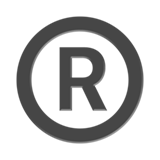}\\\midrule
        Total & 62.9M & 5.53\% & \emoji{rocket_1f680} \emoji{white-heavy-check-mark_2705} \emoji{upwards-black-arrow_2b06} \emoji{cloud_2601} \emoji{thumbs-up-sign_1f44d} \emoji{downwards-black-arrow_2b07} \emoji{party-popper_1f389}\emoji{palm-tree_1f334} \emoji{ticket_1f3ab} \emoji{heavy-check-mark_2714} \\\bottomrule
    \end{tabular}
    \label{tab:emojis_stats}
\end{table}

Table~\ref{tab:emojis_stats} also lists the 10 most frequent emojis on GitHub. Surprisingly, this list has no overlap with the top emojis on Twitter, which are \emoji{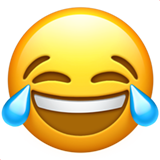} \emoji{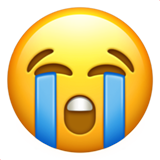} \emoji{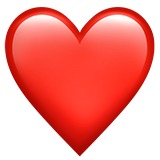} \emoji{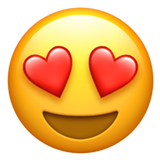} \emoji{sparkles_2728} \emoji{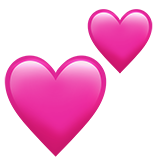} \emoji{fire_1f525} \emoji{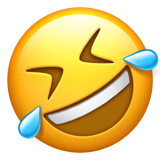} \emoji{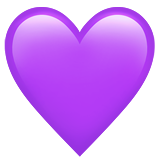} \emoji{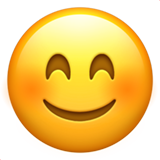}. While the top emojis on Twitter are mostly faces and hearts, the most popular emojis on GitHub, such as \emoji{rocket_1f680}, \emoji{white-heavy-check-mark_2705}, \emoji{upwards-black-arrow_2b06}, are directly related to work. For example, \emoji{rocket_1f680} is commonly used to represent the action of \textit{deploying}, \textit{launching}, or \textit{shipping}~\cite{gitmoji}, 
while \emoji{ticket_1f3ab} is a \textit{ticket} which could be used in the contexts of issue tracking and technical support. This implies that emojis are indeed used for multiple purposes on GitHub.

Different choices of emojis are also observed among different types of posts. The \emoji{thumbs-up-sign_1f44d} emoji is quite popular in comments, especially in issue comments and review comments, likely expressing positive responses to another user. The \emoji{rocket_1f680} and \emoji{party-popper_1f389} emojis are mostly used in pull request comments, pull requests, and issue comments. \emoji{police-cars-revolving-light_1f6a8} is the most used emoji in issues, usually indicating that a problem has occurred or attention is needed.

Besides these noticeable differences in Table~\ref{tab:emojis_stats}, we can still observe many emojis used on Github are clearly associated with emotions, such as facial expressions and gestures, and emotional emojis are especially popular in comments when developers respond to each other. Our analysis reveals that developers do use emojis commonly on GitHub, and they likely use emojis both for work activities and for expressing emotions. 

\subsection*{Emoji Usage by Programming Languages}
\begin{figure}[!h]
    \includegraphics[width=1\linewidth]{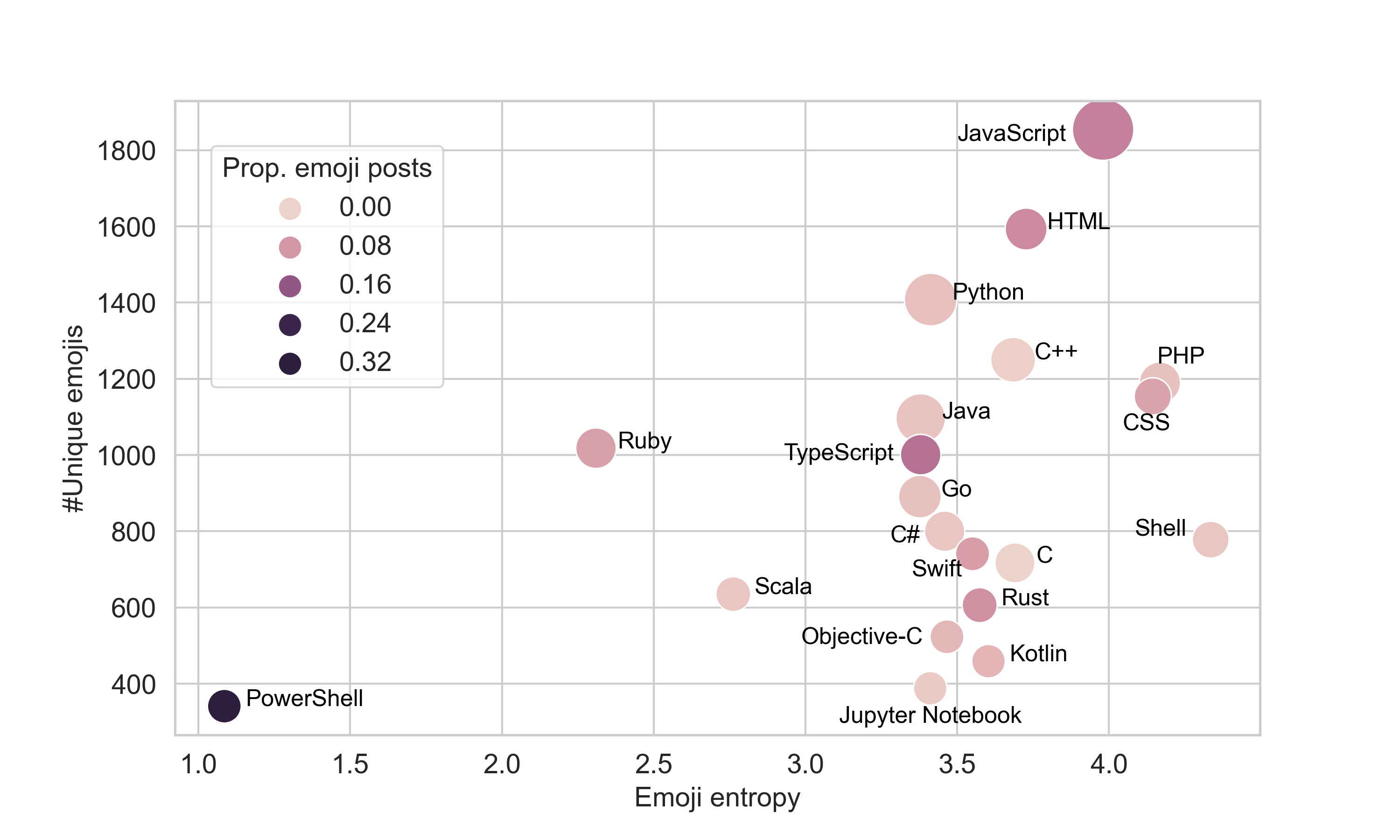}
    \caption{{\bf Emoji usage varies in repositories that have different primary programming languages.} 
    Dot: a language; size: number of repositories; color: proportion of emoji posts. JavaScript developers use a larger variety of emojis and a larger proportion of their posts contain emojis. }
    \label{fig:emoji_langs}
\end{figure}

Beyond the general popularity of emoji usage, we are interested in whether the patterns vary across different characteristics or communities of developers. Programming language is one of the most salient and noticeable identities of a developer on GitHub. We select the 20 most popular programming languages in our dataset (see \nameref{S1_appendix} for details) and plot statistics of emoji posts that identify these languages as their main languages in Fig~\ref{fig:emoji_langs}. Specifically, the size of a dot indicates the popularity of a language; the color indicates the proportion of emoji posts; and the $x$- and $y$-axis represent the entropy of emoji distribution and the number of unique emojis, two measures of the diversity of emojis used by users of that language. A higher value of entropy~\cite{shannon2001mathematical} indicates a more evenly distributed usage across all types of emojis.

Fig~\ref{fig:emoji_langs} demonstrates interesting differences among programming languages. Web programming languages, such as JavaScript, HTML, TypeScript, and CSS, are among the languages with the highest proportion of emoji posts. 
PowerShell stands as an outlier, likely because it attracts a small group of developers who prefer a small set of emojis. The differences in emoji usage across programming languages either attributes to different norms in particular developer communities or to different personalities of programmers \cite{bazelli2013personality}. Either way, they might not be related to the working status or emotions of the developers, and one should control for 
programming languages when analyzing the relation of emoji usage and work-related status and outcomes.

\section*{Emoji Usage is Related to Working Status}\label{sec:regression}

The previous analysis reveals that emojis are commonly used by developers in work-related communications. What we do not know is \textbf{\textit{whether the use of emojis is related to the working status of developers}}, especially whether the use of emotional emojis are related to working status. If the use of (emotional) emojis is unrelated to working status, then it would be less convincing to use emojis as sensors of emotions at work or as predictors for future status. Instead, if the use of (emotional) emojis can reflect the current working status of developers, we may expect that they may also have some predictive power for the future working status of developers. In this section, we answer this question through a controlled regression analysis. 

We select developers who have used at least one emoji in posts during the year 2018 and denote them as \textit{emoji users}. There are 264,808 such developers. Surprisingly, they contribute to 52.78\% of all posts in this single year, indicating that emoji usage may be related to their activeness in communication. To control for this bias, we match these emoji users with a random sample of 264,808 non-emoji users, developers who had at least one post but did not use any emoji in 2018. We collect all available work-related activities of both emoji users and non-emoji users in 2018.

\subsection*{Working Status Measurements}

We propose five categories of variables that reflect a developer's working status: 

\noindent $\bullet$ \textbf{D1: Activity level}, which measures the aggregated statistics of a developer's activities, posts, working days, working hours (hours with an activity logged), working sessions (defined as consecutive activities with gaps less than 2 hours), and off-segments (defined as longest breaks or idle periods within every 24 hours). 

\noindent $\bullet$ \textbf{D2: Trend of activities}, 
which measures the direction of linear trend (increasing, decreasing, or stable) of one's activity levels by month. 

\noindent $\bullet$ \textbf{D3: Type of activities}, 
based on activity types included in our dataset. We select a subset of activities, including \textit{push}, \textit{pull request}, \textit{issue}, and all activity events related to \textit{comments} 
and count their proportions among all work activities and the sessions/days containing these types of activities. 

\noindent $\bullet$ \textbf{D4: Type of posts}, proportions of each type of posts among all posts, or work-related communications among developers. 

\noindent $\bullet$ \textbf{D5: Time management}. Time management may reflect the work-life balance of a developer and have a direct relationship with their mental health status. Because the dataset does not provide time zones of users, we measure the entropy of one's working hours over the 24 hours of a day and working days over the 7 days of a week, respectively. A higher entropy of hours or days indicates more spread-out working hours or days, while a lower entropy could indicate clear boundaries between work and life.

Note that all the variables are measured based on the developer's activities we observe on GitHub, and they don't reflect their work activities offline, in private repositories, or on other platforms. 
We are interested in understanding whether the work-related variables extracted are related to emoji usage at all, both individually and collectively. Having potential confounds controlled, if the work-related variables are still significant factors of the use of emojis, then we may confirm that emojis can be used as sensors of the working status of individual developers. 

\begin{figure}[!h]
\includegraphics[width=1\linewidth]{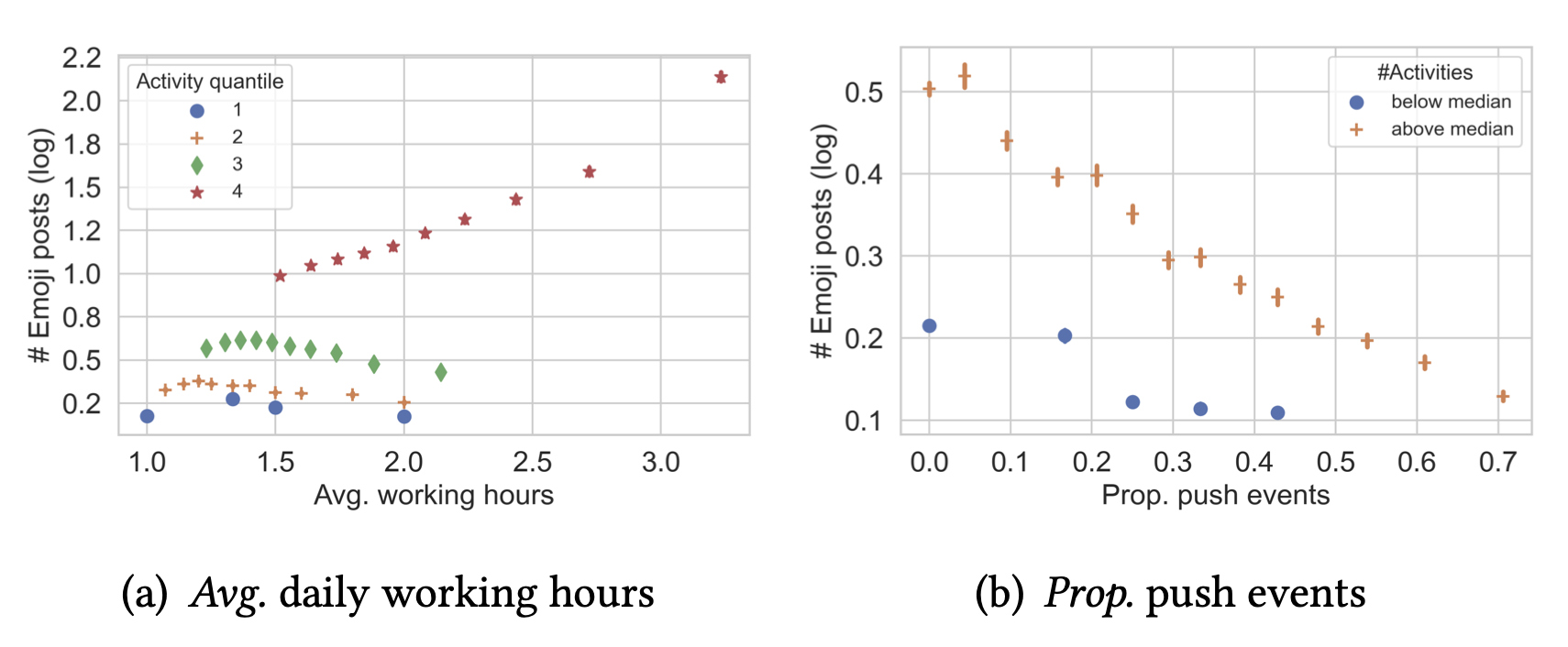}
\caption{{\bf Working status of a developer is related to the number of emoji posts.}
(a): \textit{Avg.} daily working hours. (b): \textit{Prop.} push events.}
\label{fig:corr}
\end{figure}

We find that measures of emoji usage (both the number and the proportion of emoji posts a developer wrote in the same year) present interesting correlations with many of these working status measures. We show two examples in Fig~\ref{fig:corr}.  
In Fig~\ref{fig:corr} (a), we partition developers into four categories based on the quantiles of their activity levels (\#activities).  We find that developers who have longer working hours tend to write more emoji posts -- but only for those in the most active quantile (4). Interestingly, we don't see a similar trend for the less active developers (quantiles 1-3).  Fig~\ref{fig:corr} (b) also suggests different effects of the proportion of push events on the emoji posts among active users and inactive users. 

To reveal how emoji usage is related to the selected variables jointly, 
we conduct a series of linear regressions (OLS) with the number of emoji posts and the proportion of emoji posts of a developer as outcome variables. Specifically, to evaluate if emotional emojis alone are related to working status, we also conduct an OLS regression with the proportion of emotional emojis as an outcome variable. We control for ``demographical'' variables of a developer that may confound the relation between emoji usage and working status. Besides a developer's primary programming language (which has been shown to be relevant), we also control for their tenure on GitHub (platform age).

\subsection*{Regression Analysis}\label{subsec:regression}

Combining all the features, we have 64 candidate variables to characterize the working status of a developer as well as their online ``demographics.'' A regression with all of the variables would lead to multicollinearity and deteriorate the results. We leverage the Variance Inflation Factor (VIF)~\cite{james2013introduction}, which detects whether an explanatory variable is highly collinear with other variables, resulting in large standard errors for the coefficient estimates. 
Using the \texttt{variance\_inflation\_factor} API in the \texttt{statsmodels} package~\cite{statsmodels}, we repeatedly run regressions and remove the explanatory variable with the largest VIF until all variables' VIF are below 5, a commonly used cut-off~\cite{sheather2009modern}. 
This leaves us with 45 variables in total. Compared with alternative methods such as PCA, such a variable selection method ensures the explainability of the model. 

With the selected independent variables, we add them into the OLS models one category at a time. 
We report two regressions in Table~\ref{tab:regression_general_shortened} and the full results in Table S1, S2, and S3 in \nameref{S1_appendix}. Configuration (5) includes variables from D1-D5, and configuration (6) adds the ``demographics'': platform age and programming languages.

\begin{table}[!ht]
\centering
\caption{{\bf OLS Regressions of Emoji Usage on Working Status.} Multiple working status measures are significantly related to emoji usage. }
\begin{tabular}{lcccccc}
\toprule
\multicolumn{3}{c}{Dependent Variable: Number of Emoji Posts (Log Scale)}\\\midrule
                           &  (5) & (6)  \\
\midrule
const	&	2.4833***	&	2.0552***	\\
\textit{avg.} working hours	&	0.0240***	&	0.0334***	\\
\textit{avg.} length of working sessions (in hour)	&	0.0003**	&	0.0002*	\\
length of off segment (thres. = 4, 6, 32, 64, 128)	&	$\triangle$***	&	$\triangle$***	\\
length of off segment (threshold = 256)	&	0.0104**	&	0.0145***	\\
monthly trend of \#working days	&	-0.0101***	&	-0.0069***	\\
monthly trend of \textit{avg.} working hours	&	0.0098***	&	0.0117***	\\
monthly trend of \textit{avg. len.} of working sessions	&	-0.0001	&	0.0019	\\
monthly trend of \#working sessions	( $>$1 event) &	-0.0077***	&	-0.0068***	\\
monthly trend of \#posts	&	0.0105***	&	0.0113***	\\
\textit{prop.} pull request events	&	0.2979***	&	0.2079***	\\
\textit{prop.} push events	&	-0.3261***	&	-0.3165***	\\
\textit{prop.} working sessions with pull requests 	&	0.4902***	&	0.5022***	\\
\textit{prop.} working days with comment events	&	0.3839***	&	0.4069***	\\
\textit{prop.} working days with issue events	&	0.4348***	&	0.3983***	\\
\textit{prop.} pull requests	&	-0.1428***	&	-0.1269***	\\
\textit{prop.} issue comments	&	0.0223***	&	0.0176***	\\
\textit{prop.} pull request comments	&	0.5859***	&	0.5622***	\\
\textit{prop.} pull request review comments	&	0.6380***	&	0.6093***	\\
\textit{prop.} commit comments	&	0.0355***	&	0.0559***	\\
entropy of weekdays	&	0.3188***	&	0.3030***	\\
platform age (log scale)                           &                        & 0.0373***                \\
programming languages                &                        & $\diamond$                \\
\midrule				
\# Obs.	&	529616	&	529616	\\
R-squared	&	0.4704	&	0.4845	\\
Adj. R-squared	&	0.4704	&	0.4846	\\
					
\bottomrule
\end{tabular}
\begin{flushleft} \textit{Notes}: Significant at the: *** 1\%, ** 5\%, or * 10\% level;\\
$\triangle$: All coefficients are negative. $\diamond$: All languages are significant except for Objective C. See \nameref{S1_appendix} for details.
\end{flushleft}
\label{tab:regression_general_shortened}
\end{table}

Table~\ref{tab:regression_general_shortened} shows that even after controlling for programming languages and platform age, most of the working status related variables are still significant in the regression with the number of emoji posts as the dependent variable. The signs, scales, and significance of the coefficients are mostly robust. Similar observations can be made using the proportion of emoji posts as the dependent variable, which we omit for the sake of space.  The regression results are promising, suggesting that emoji usage is highly reflecting working status of individual developers rather than community norms. 

We are also able to examine the effect of certain variables more closely from the table. 
The activity level of a developer, such as the average working hours (on working days) or the length of working sessions, is with a positive coefficient with the number of emoji posts. This is consistent with the plot in Fig~\ref{fig:corr} (a). 

Most of the off-segments lengths (i.e., longest idles) show a negative sign in the regression of the number of emoji posts, which is consistent with the finding that a higher activity level indicates using more emojis. The off-segment length with the highest thresholds (256, meaning the longest idle length in at least 256 days in 2018) is an exception, which suggests that the relationship could be non-linear. 

Developers with an increasing trend of \textit{avg.} working hours, or \#posts are likely to have more emoji posts (as well as a higher proportion of emoji posts), while those with an increasing trend of \#working days compose fewer (but a higher proportion of emoji posts, suggesting much fewer posts overall). An increasing activity level or workload might be associated with either passion or stress at work and a decreasing trend might be an indicator of burnout. 

Developers with more push activities compose fewer emoji posts, while those with more activities in issues, pull requests, and comments compose more emoji posts. This result is reasonable as with other variables controlled, developers with more push events could pay more attention to coding rather than communication. 

Developers with more comments compose more emoji posts, suggesting that emojis can mean appreciation and encouragement from the collaborators. 

The entropy of working days over the week has a positive effect on emoji posts (but a negative effect on their proportion). Note that a higher entropy indicates a more even spread-out of working activities over the week, which may be linked to a higher passion or stress that drives them to work during the weekends. 

Finally, we note that the control variables, platform age, and most programming languages are significant factors in the regressions of emoji usage. This is consistent with our observations in Fig~\ref{fig:emoji_langs}, and the positive effect of platform age indicates that developers are gradually establishing the norm of using emojis on GitHub.

As not all emojis are used to convey emotions, it is perhaps not surprising that work-related emojis are related to work activities of developers. To illustrate the relation between emotions expressed through emojis and the working status of developers, we conduct separate OLS regressions with the proportion of emotional emojis as the outcome variable. Specifically, we extract the keywords describing each emoji in the Unicode Standard, get their emotion scores through LIWC, which are \textit{positive emotion}, \textit{negative emotion}, \textit{anger}, \textit{anxiety}, and \textit{sadness}. An emoji is identified as emotional if it has a non-zero emotion score on one or more of the five dimensions. We then compute the proportion of emotional emojis among all emojis used by a developer. As shown in Table S3 in \nameref{S1_appendix}, most of the working status variables are still significant when the proportion of emotional emojis is the response variable. This result shows that the intensity of emotions expressed via emojis is an efficient indicator of the current working status of developers.

\subsection*{Emoji Usage Indicates Future Work Engagement}

The preceding results have demonstrated that emoji usage is related to the working status of developers in the same year. Can emoji usage indicate the future status of developers? To validate this, we select two typical measures of working status, i.e., \textit{avg.} daily working hours and the number of working days, and we compare these measures in the next year (2019) between developers who used emojis in the first year (2018) and those who did not. Note that only developers who wrote at least one post in 2018 and were still active in 2019 are included in this analysis. In 2019, emoji users worked for 62.39 days on average, which is significantly higher than 23.13 days of non-emoji users ($p \ll 0.001$, Welch two-sample t-test). The difference of average daily working hours between the emoji users (1.71 hours) and non-emoji users (1.45 hours) is also significant ($p \ll 0.001$, Welch two-sample t-test). Such results show that emoji users are more engaged in the coming year. This anecdotal evidence implies that we may use emoji usage as features to predict certain future work outcomes of developers in a more systematic way.

\section*{Emoji Usage Predicts Future Dropout}\label{sec:prediction}
The insights drawn from preceding analysis suggest that emoji usage may be useful to predict future working status of developers. We are particularly interested in using emoji usage as a lens to watch for outcomes related to the emotional and mental health related well-being of workers, such as passion \cite{vallerand2003passions}, depression~\cite{Woo2011-hs}, and burnouts~\cite{github2020spotlighht}. These outcomes, however, are hard to be accurately measured at scale.  We therefore choose a more explicit prediction target: whether a developer will conduct \textit{no} activity on GitHub (i.e., whether they will drop out from the platform) in the coming year, which is often related to the emotions and mental health~\cite{wilcox2004early, bask2013burned}. In this section, we describe the emoji usage of a developer at finer granularities and answer the third research question, \textbf{\textit{whether emoji usage can be used to predict the dropout of developers}}.  In specific, we extract features of emoji usage from 2018 and use them to predict whether a developer active in 2018 would drop out from the work activities on GitHub (defined as zero working days or daily working hours) in 2019. 

Since emoji usage is highly related to activity level, and both are related to programming languages, we need to carefully control for such confounding factors in the prediction task. In below, we first discuss how we control for the two confounding factors before setting up the prediction task. 

\subsection*{Developer Matching}

To find good measurements of the activity level, we draw insights from labor economics where researchers have long studied the intensive and extensive margin of labor supply~\cite{blundell2013extensive}. Extensive margin measures whether or not a unit is working, corresponding to the working days of a developer, while intensive margin measures the intensity of one's work, corresponding to the average working hours. We adopt both the number of working days and the average working hour on a working day as the measures of the activity level of a developer in both 2018 and 2019.  The measures in 2018 are used to control for the activeness of developers, and the measures in 2019 are used to identify the prediction outcome. 

We define a target group of developers with a certain activity level in the previous year. Specifically, we rank all developers by their activity level (measured by either \#working days or \textit{avg.} working hour) in 2018 and select the $k$\% \textit{most active} developers into the group. For average working hours, we choose $k$ from 5, 10, 20, 30, 40, 50, and 60, since over 40\% of developers share the same lowest level of this measure. For the same reason, we select $k$s of 5, 10, 20, 30, 40, 50, 60, and 70 for \#working days. Through this, we obtain multiple (overlapping) target groups at comparable activity levels. The smaller $k$, the more active developers in the target group.  

As discussed in previous sections, the primary programming language of a developer is clearly a confounder to their emoji usage and working status. In Fig S2 in \nameref{S1_appendix} we further demonstrate the relation between the working status outcome and the programming languages. 
To control for this confounding factor, we match \textit{emoji users} with a set of \textit{non-emoji users} who have a similar distribution of programming languages within the same activity level. For instance, if there are $m$ JavaScript emoji-users at one activity level, we randomly sample $m$ JavaScript users from the $n$ JavaScript non-emoji users within the same activity level. $n$ is usually much larger than $m$, but in the case that $n<m$, we select all $n$ JavaScript users and randomly sample $m-n$ users from those who are not labeled with a primary programming language. 

Among all emoji usage patterns, the most basic one is whether a user uses emoji at all. Before introducing more complex emoji-usage patterns, we show that this simplest distinction of emoji users versus non-emoji users is already predictive of dropout. 

Controlling for programming languages, non-emoji users have a \textbf{3 times higher} ratio of dropouts in the following year than emoji users. Including all activity levels, 13.0\% of emoji users and 49.2\% of non-emoji users drop out in 2019 ($p \ll 0.001$, Welch two-sample t-test); among top 5\% developers by the number of working days in 2018, 1.4\% of emoji users and 4.5\% of non-emoji users drop out in 2019 ($p \ll 0.001$, Welch two-sample t-test).   

\subsection*{Dropout Prediction for Emoji Users}

The above comparisons show that the mere fact of using emoji already signals less likelihood of dropout. Yet, beyond this basic one, there are many other emoji usage patterns. To systematically study whether they help predict the dropout of developers, we set up a prediction task where we extract the emoji-usage patterns and see whether they predict a user will drop out in the next year. 
For this prediction task, we include only the emoji users since the emoji-usage features for non-emoji users are either empty or zero. Among the emoji users at each activity level, those who have no activity in 2019 are labeled as positive examples; we then sample the same number of \textit{emoji users} from the rest of the target user group, those who do not drop out in 2019, as negative examples. We use the same sampling procedure above to match the primary programming languages of the negative examples.  

This matching procedure ensures the balance of positive and negative samples within each activity level in 2018 and that the two groups share a similar distribution of programming languages. For robustness, we repeat the sampling processes 30 times and construct 30 datasets for each activity level. 

We use four advanced machine learning models, Logistic Regression, Support Vector Machine (SVM),  Gradient Boosting Decision Tree (GBDT), and Multi-layer Perceptron classifier (MLP), for this binary classification task, all implemented with scikit-learn~\cite{scikit-learn}. We use accuracy as the evaluation metric. As all the datasets are balanced, the baseline accuracy is 0.5. We use an 80-20 train-test split for each dataset (target group) and tune hyper-parameters on training with 5-fold cross-validation.

Note that our goal is to verify the predictive power of emojis instead of optimizing the performance of dropout prediction. Therefore we intentionally exclude non-emoji-based features (such as the working status measures in the previous section) and more complex machine learning models so that the whole process is highly explainable. 

\subsection*{Emoji Usage Features}
We summarize the emoji-usage features used in the prediction task. First, we follow the design of the regression in the previous section and measure the number and proportion of emoji posts of a developer in 2018, which are shown to be indicative of the working status in the same time period. We count the number of days when emojis are used as well as the total number of emojis used by the developer.

To describe the diversity of emojis used by a developer, we count the number of unique emojis they have used in 2018 and calculate the entropy of the frequency distribution over unique emojis.

Usage of individual emojis is also considered. From more than two thousand kinds of emojis used on GitHub (2,699 in our dataset), we select a subset of representative ones to ensure explainability and to avoid overfitting. Specifically, we choose emojis from three aspects. First, we select 30 most used emojis by all developers, representing the most popular emojis on GitHub. Second, we select the emojis that are used significantly more frequently on GitHub than on Twitter. That is, we select the 30 emojis with the highest proportions on GitHub compared to Twitter, ranked by their $z$- statistics (one-tail z-test), and denote them as \textit{GitHub-specific} emojis. Third, we include the 36 emojis that are reported to be related to personality traits (\textit{neuroticism}, \textit{extraversion}, and \textit{agreeableness} of the Big-Five traits)~\cite{marengo2017assessing}. In literature, 10 of the 36 emojis have been proposed as surrogates to text-based items to assess symptoms of depression~\cite{marengo2019development}. Removing the duplicates of the three groups, we finally obtain 74 unique emojis.

To quantify the emotional status of a developer, we measure the emotions conveyed by all emojis used in their posts in 2018. Specifically, we extract the keywords describing each emoji in the Unicode standard, get their LIWC~\cite{pennebaker2015development} emotion scores, and aggregate the scores to measure multi-dimensional emotions of that emoji. We first include the proportion of all emojis that have a non-zero emotion score for each user. By averaging the emotion scores of all emojis by a user, we obtain five scores, i.e., \textit{positive emotion}, \textit{negative emotion}, \textit{anger}, \textit{anxiety}, and \textit{sadness}. This provides a 5-dimensional description of the developer's emotional status. Additionally, we include the EMOJI-D score proposed in~\cite{marengo2019development} to measure the depression level of a developer by aggregating the usage of the corresponding 10 emojis.  

\subsection*{Prediction Results}

\begin{figure}[!h]
\includegraphics[width=1\linewidth]{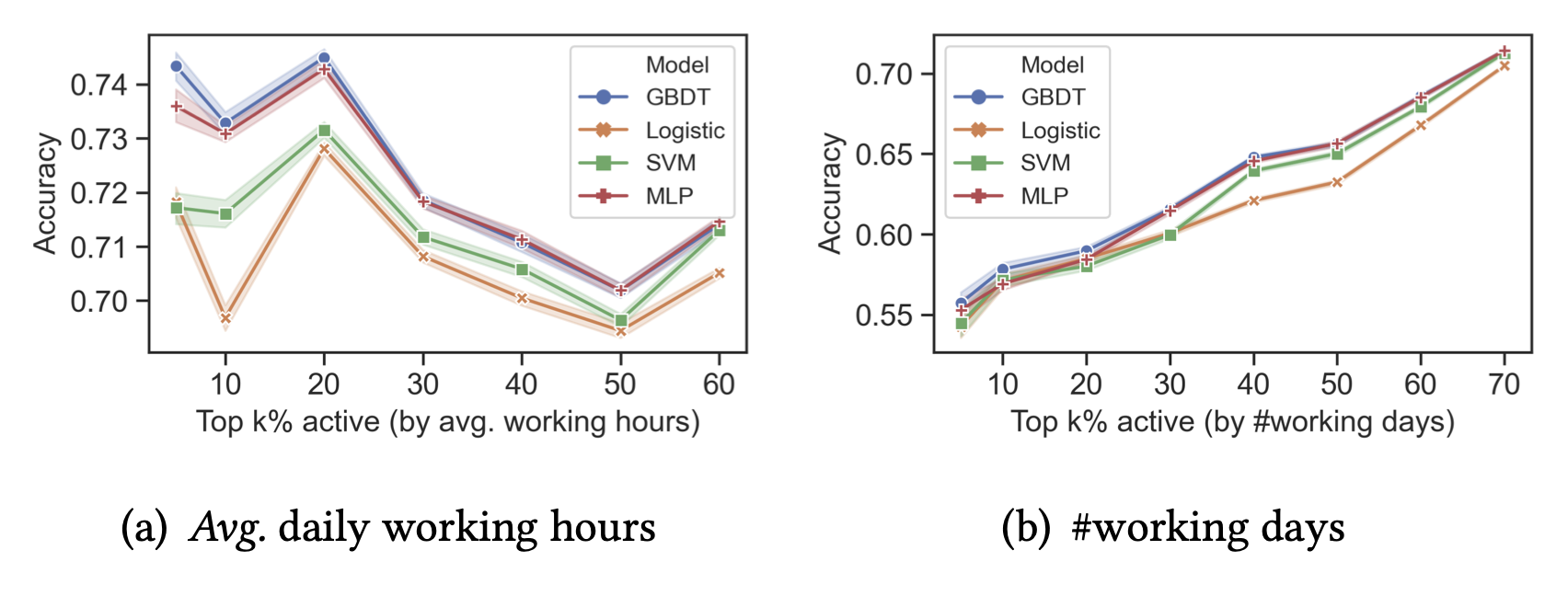}
\caption{{\bf Accuracy of dropout predictions for emoji users is above 0.7 when including all activity levels and as high as 0.75 for those who work longer hours per working day.}  
Baseline is 0.5 for all activity levels and is omitted. (a): \textit{Avg.} daily working hours. (b): \#working days.} 
\label{fig:pred}
\end{figure}

For all four models, Fig~\ref{fig:pred} reports the mean accuracy and 95\% confidence intervals over all 30 datasets at each controlled activity level. For all activity levels measured by average working hours, all four prediction models obtain a decent accuracy (mostly above 0.7). For example, when $k=5$, the accuracy is 0.718 for Logistic Regression, 0.717 for SVM, 0.736 for MLP, and 0.743 for GBDT. This level of prediction accuracy is rather impressive considering the difficulty of dropout prediction and that only emoji usage is used as features. 

Controlling for activity levels by the number of working days, the prediction accuracy approaches the same level when $k$ is large (including less active developers), but it's lower when $k$ is small (more active developers). This is understandable as developers who worked for the most number of days in the previous year are much less likely to dropout in the coming year, and those who do are much harder to predict. On the other hand, developers who worked for the longest hours on the days they work might not have worked for many days. We further interpret the prediction performance for the most active developers (when $k$ is small). A precision-recall curve is shown in Fig S1 in \nameref{S1_appendix}. It suggests that even though the overall prediction accuracy is not high among the top 5\% of active developers (by working days), the precision for those predicted the highest dropout risk is as high as 0.73. Considering the nature of the prediction task, this level of accuracy is sufficient for taking interventions on the developers with highest dropout risks. 

\begin{figure}[!h]
\includegraphics[width=1\linewidth]{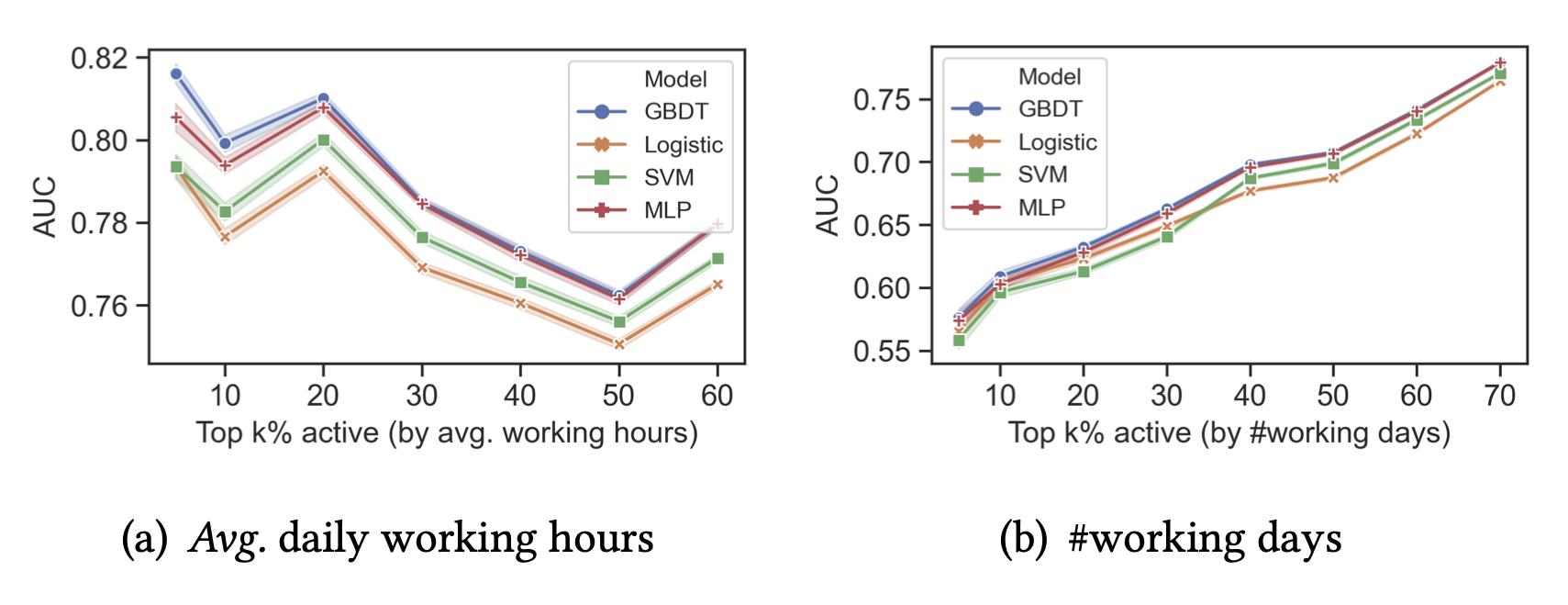}
\caption{{\bf AUC of dropout predictions for emoji users is above 0.78 when including all activity levels and as high as 0.82 for those who work longer hours per working day. }
 (a): \textit{Avg.} daily working hours. (b): \#working days.}
\label{fig:pred_auc}
\end{figure}

The performance of the prediction models could also be measured by AUC (i.e., Area under the ROC Curve). Fig~\ref{fig:pred_auc} reports the mean AUC and 95\% confidence intervals over the 30 datasets at each controlled activity level for the four models. All the models obtain a high AUC for all activity levels measured by average working hours, mostly above 0.75. For example, when $k=5$, the AUC is 0.793 for Logistic Regression, 0.794 for SVM, 0.805 for MLP, and 0.816 for GBDT. 

As expected, the non-linear GBDT models outperform SVMs and Logistic Regressions. GBDT models also outperform MLPs for most activity levels. Overall, the differences between the four standard machine learning models are marginal.

\subsection*{Interpretations}

\begin{figure}[!h]
    \includegraphics[width=\linewidth]{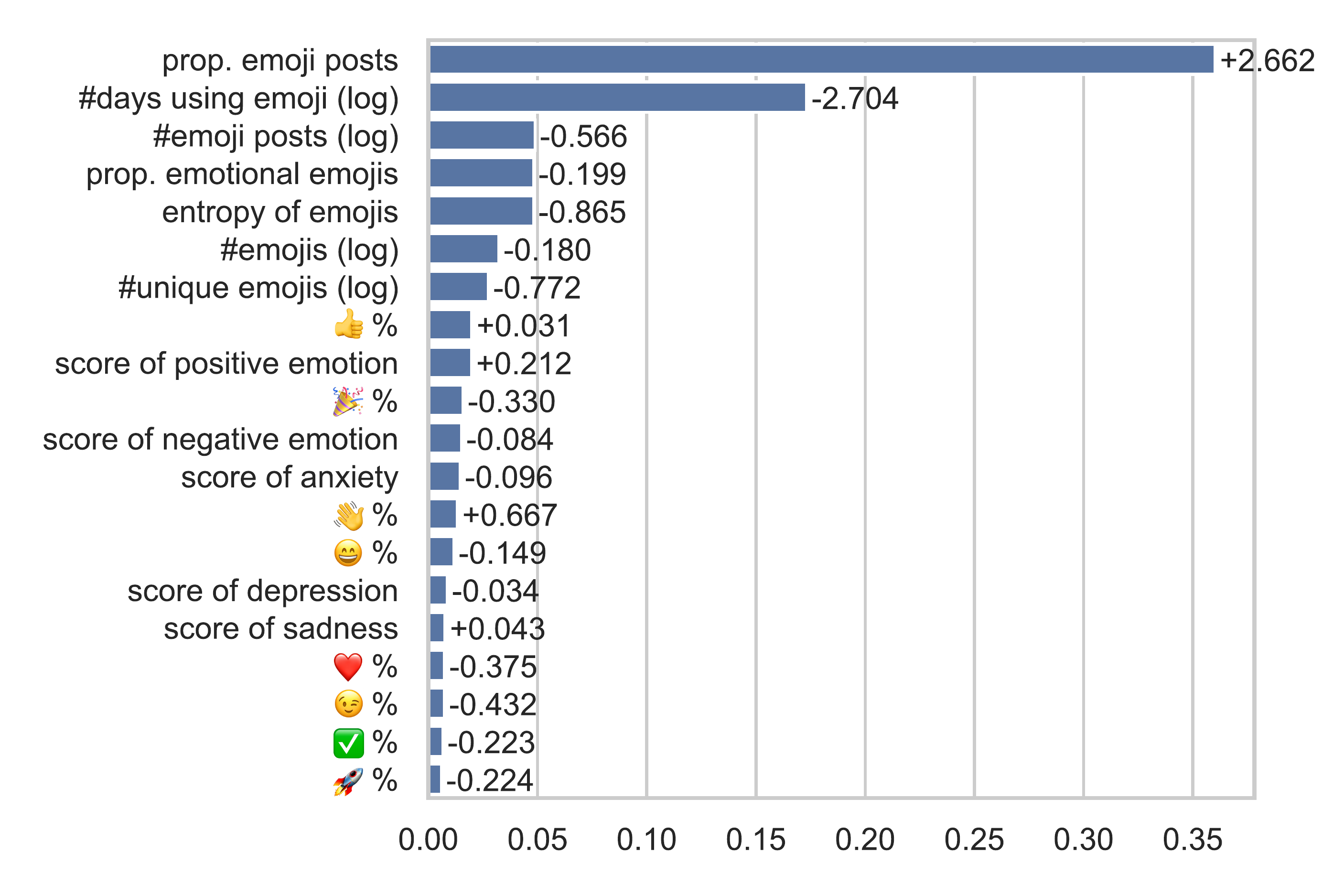}
    \caption{ {\bf Importance of top features (10\% active users by \textit{avg.} working hours). Proportion of emoji posts, emotional scores, and affection-related emojis are important predictors. } Bars: importance scores from GBDT; signed values: coefficients from Logistic Regression.}
    \label{fig:avg_hours_daily_feature}
\end{figure}

\begin{figure}[!h]
\includegraphics[width=1\linewidth]{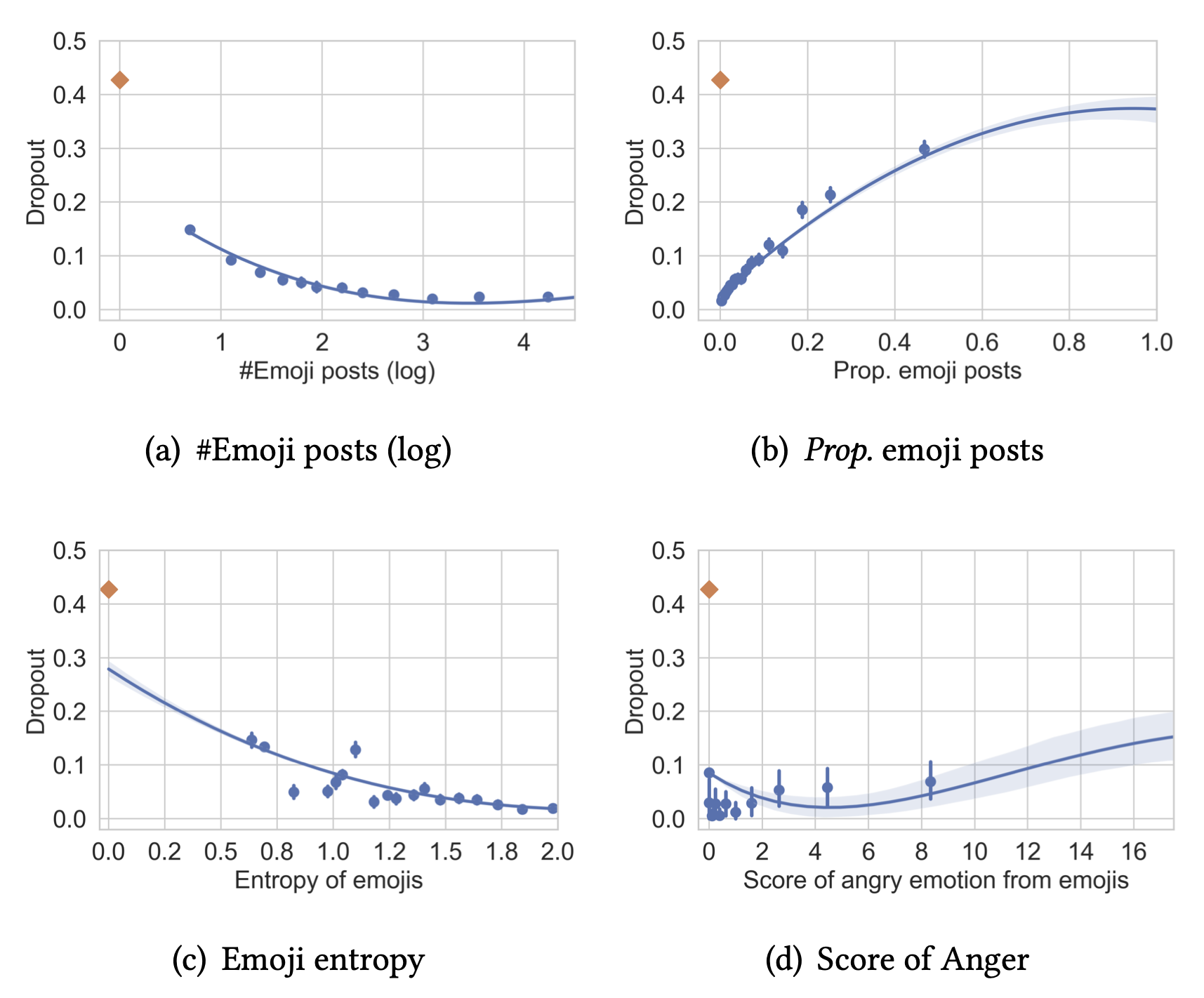}
\caption{{\bf Relation between selected emoji usage features and dropout risk (10\% active users by \textit{avg.} working hours).} Blue dots: bins of emoji users; orange diamond: non-emoji users. (a): \#Emoji posts (log). (b): \textit{Prop.} emoji posts. (c): Emoji entropy. (d): Score of Anger emotion. }
\label{fig:feature_interpret}
\end{figure}

The results above demonstrate that future dropout of developers can be effectively predicted using emoji usage features alone. 
To further understand and explain the role of emojis in predicting dropouts, we look into the importance of features provided by the GBDT model and the coefficients of features obtained by the Logistic Regression model, using one configuration with top 10\% ($k=10$) developers by average working hours as an example, as shown in Fig~\ref{fig:avg_hours_daily_feature}. Features with high importance scores in GBDT have relatively high predictive power for dropouts. The sign of LR coefficients indicates the linear direction of the predictive power, although the actual effect might be non-linear. We therefore plot the relation between the likelihood of dropout and features of interest for a closer look, as shown in Fig~\ref{fig:feature_interpret}. Recalling that we use only emoji users in the prediction task, for the plots, we include both emoji users and non-emoji users at the matched activity level ($k=10$) to understand the role of emojis in a comprehensive way.

Fig~\ref{fig:avg_hours_daily_feature} shows that features describing the general usage of emojis, especially the proportion of emoji posts, appear to be the most predictive. All 5 dimensions of emotion scores and the proportion of emotional emojis are important in the prediction task. In addition, the proportion of some individual emojis are ranked high by GBDT, mostly those expressing affections (e.g., \emoji{thumbs-up-sign_1f44d}, \emoji{party-popper_1f389}, smiley faces, and the heart \emoji{heavy-black-heart_2764}). This suggests that the emotional use of emojis is indeed a viable indicator of dropout. The fact that the \textit{general usage of emojis} (regardless of emotional or not) remains important features indicates that there may exist other purposes of emoji use - beyond expression emotions - that are also indicative of developer dropout. Untangling these factors would be more difficult than measuring emotions, which requires fine-grained annotations of the purposes of individual emoji usecases.

Fig~\ref{fig:avg_hours_daily_feature} also shows the coefficients of the selected features in Logistic Regression. A positive coefficient suggests that the feature indicates a higher risk of dropout with other features controlled, and a negative coefficient indicates a lower risk of dropout. It is interesting that the proportion of emoji posts and the number of emoji posts (as well as the number of emojis) have different signs. Verified at a finer granularity in Fig~\ref{fig:feature_interpret} (a) and Fig~\ref{fig:feature_interpret} (b), we see that the correlation trends are not linear. The result suggests that if developers use emojis in their posts at all, they are much less likely to dropout, and the risk further decreases when they compose more (but a reasonable number of) emoji posts. However, when a high \textit{proportion} of their posts contain emojis, the relation points to the other direction. These users are possibly obsessive about emojis, demonstrating potential signals of obsessive passion \cite{lavigne2012passion}.  

The diversity of emoji usage, both the number of unique emojis and the entropy of emojis, show a negative effect on the risk of dropout. Fig~\ref{fig:feature_interpret}(c) shows a descending trend of dropout risk with higher entropy of emojis. A diverse usage of emojis may indicate that a user is in good emotional status (e.g., curiosity \cite{Kashdan2006-ba}) to explore and indicates a lower risk of obsession or burnouts.

Emotions expressed through emojis also show intriguing relations to dropouts. Most of the emotion scores, \textit{i.e.}, the \textit{positive} emotion, the \textit{negative} emotion, \textit{sadness}, and \textit{anxiety}, have negative coefficients, while \textit{anger} shows a positive effect. We plot the relations of these features to dropout ratio for a closer look. Similar to what Fig~\ref{fig:feature_interpret} (d) presents, the relations are actually not linear. A zero value of these emotions seems to be a clear outlier with a significantly higher dropout rate. This indicates that expressing these emotions -- even if they are negative -- using emojis at all reduces dropout risk. Leaving out this outlier, we can observe positive correlations of these emotions and mental health-related dimensions to dropouts.  Among them, the slope of the \textit{anger} emotion is the steepest, indicating that \textit{anger} is a sensitive signal of future dropouts. This result conforms with previous literature~\cite{ersoy2009relations, Elfenbein2007-no} that anger levels differ according to burnout levels, suggesting that anger-related emojis might be used to detect burnouts of developers. 

Interesting findings could also be derived from the use of individual emojis. The \emoji{waving-hand-sign_1f44b} emoji is positively correlated with dropouts, as it is a common gesture for leaving. The negative coefficients of affection and celebration-related emojis (e.g., \emoji{heavy-black-heart_2764},  \emoji{party-popper_1f389}, and  \emoji{smiling-face-with-open-mouth-and-smiling-eyes_1f604}) are easier to interpret. The findings are consistent with psychology and organization science literature that affections and personal achievements are highly correlated to work outcomes \cite{Staw1994-lx, Thoresen2003-wt}. It is harder to interpret why \emoji{thumbs-up-sign_1f44d} has a marginal positive sign, likely due to a non-linear relation again. 

So far we have explored the role of emojis in dropout prediction through the relations between specific features and the dropout risk. To further understand why emoji usage is predictive of future dropout, we propose two possible explanations through additional correlational analysis. One is that emoji usage patterns can reflect certain intrinsic characteristics of workers, that they have better emotional intelligence at work, such as the willingness to express and regulate emotions in working scenarios, which is known to be related to their working performance~\cite{Rafaeli1987-tt}. As such intrinsic characteristics hold for a long term, we can expect that the emoji usage in the previous year is correlated with emotions expressed in the next year.

\begin{figure}[!h]
\includegraphics[width=1\linewidth]{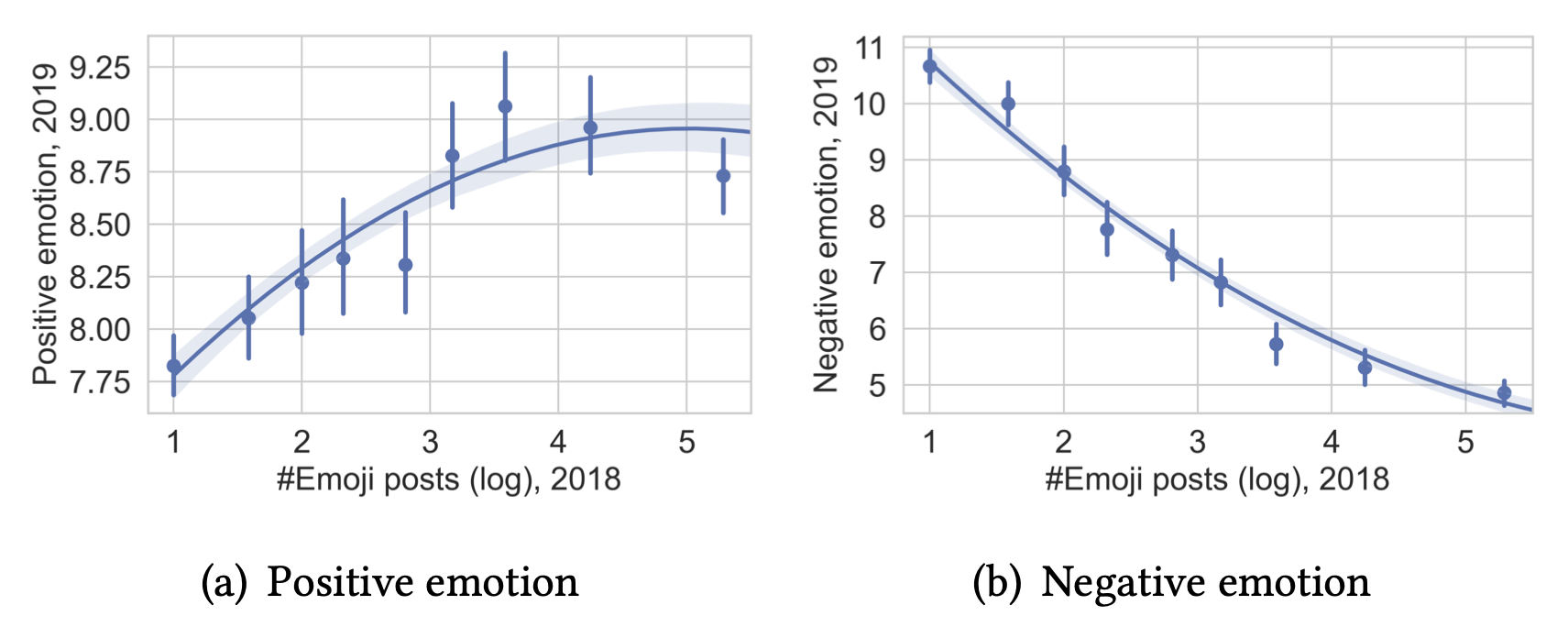}
\caption{{\bf Emotion expressed via emojis in 2019 is related to the number of emoji posts in 2018.}
(a): Positive emotion. (b): Negative emotion.}
\label{fig:emoji_post_log2_18_vs_emotion_19}
\end{figure}

We measure the positive and negative emotional scores of emojis used by a developer and compare the emotions in the two consecutive years in Table~\ref{tab:emotions}. In general, developers with at least one post in 2018 express more positive emotions and less negative emotions in 2019. However, those who start to use emojis in 2019 express more negative emotions, while those who already used emojis in 2018 are much happier in 2019. Fig~\ref{fig:emoji_post_log2_18_vs_emotion_19} presents the relation between the positive/negative emotions in 2019 and their emoji usage in 2018. Developers tend to express more positive emotions and less negative emotions if they use more emojis in the previous year. Such results support our hypothesis and also suggest that emoji users have a more positive emotional status.

\begin{table}[]
\centering
\caption{{\bf Positive and negative emotions expressed via emojis in 2018 and 2019.} In general, the population becomes more positive and less negative in 2019. Those who used any emoji in 2018 are more positive ($p<0.001$, Welch two-sample t-test) and less negative ($p<0.001$, Welch two-sample t-test) in 2019 than those who didn't use an emoji in 2018.}
\begin{tabular}{lcccc}
\toprule
\multicolumn{1}{c}{\begin{tabular}[c]{@{}c@{}}Developers in 2018\\ (\#post\textgreater{}0)\end{tabular}} & \begin{tabular}[c]{@{}c@{}}Positive\\ (2018)\end{tabular} & \begin{tabular}[c]{@{}c@{}}Negative\\ (2018)\end{tabular} & \begin{tabular}[c]{@{}c@{}}Positive\\ (2019)\end{tabular} & \begin{tabular}[c]{@{}c@{}}Negative\\ (2019)\end{tabular} \\
\midrule
Emoji users      & 7.54    & 9.45    & 8.27    & 8.51    \\
Non-emoji users      & \textit{N/A}     & \textit{N/A}     & 7.61    & 9.55    \\
All       & 7.54    & 9.45    & 7.90    & 9.09    \\
\bottomrule
\label{tab:emotions}
\end{tabular}
\end{table}

Another possible explanation is that as non-verbal cues, emojis can bring in external effects (that plain texts could not) that in turn influence the working status of developers (emoji users).  We may validate this from a ``micro'' perspective. For example, using pictographs in a post may attract more attention and therefore make the communications more efficient, or they could attract more responsive emotional supports and therefore enhance co-worker bond \cite{ren2007applying}. We select all issues opened in 2018 and their comments posted within 365 days, and find that for all issues with at least one comment, the average response time for emoji issues is 215.64 hours, significantly lower than that for issues without an emoji (420.78 hours). We also find that an emoji issue receives 0.92 emoji comments on average, while the number for non-emoji issues is only 0.08. The results are all reported in Table~\ref{tab:issue}. The higher frequency of emojis occurring in comments implies more emotional support.

To summarize, we find that emoji use in general is predictive of developer dropout. The emotional use of emojis in particular is one important indicator, although it is likely not the only viable indicator. Note that none of our results indicate any causal relation between emoji usage and work, and the above analyses certainly do not establish causal paths. We intend to provide possible explanations for why emotional emojis are related to working status and why emojis could predict worker dropout.

\begin{table}[]
\centering
\caption{{\bf Difference of responses between issues with and without emoji.} For issues with at least one comment, the time to receive the first comment for emoji issues is significantly shorter than non-emoji issues ($p<0.001$, Welch two-sample t-test). Emoji issues get significantly more comments containing at least one emoji ($p<0.001$, Welch two-sample t-test). }
\begin{tabular}{lrrr}
\toprule
  & \multicolumn{1}{c}{\begin{tabular}[c]{@{}c@{}}\#issues with \\ comments*\end{tabular}} & \multicolumn{1}{c}{\begin{tabular}[c]{@{}c@{}}\textit{Avg.} response \\ time (hour)\end{tabular}} & \multicolumn{1}{c}{\begin{tabular}[c]{@{}c@{}}\textit{Avg.} \#emoji \\ comments\end{tabular}} \\
  \midrule
Emoji issues  & 204,047  & 215.64  & 0.92  \\
Non-emoji issues & 4,662,957  & 420.78  & 0.08  \\
All issues  & 4,867,004  & 412.18  & 0.11  \\
\bottomrule
\label{tab:issue}
\end{tabular}
\begin{flushleft} \textit{Note}: *: For each issue, only the comments created within 365 days of the issue's open time are counted.
\end{flushleft}
\end{table}

\section*{Discussion}
\subsection*{Implications} 

The prediction analysis successfully demonstrates that emoji usage patterns are indicative of the risk of dropping out from online work platforms. 
Several practical implications could be derived here. First, using (some) emojis at work indicates a lower risk of dropout than not using emojis. Expressing any emotion, even negative emotions with emojis is better than expressing no emotion at all. On one hand, expressing emotion itself is a good way to handle stress~\cite{clark2004does}; on the other hand, it enables co-workers and practitioners to observe negative signals and risks early on and make timely interventions. In online work platforms where people don't express emotions face-to-face, it is important to encourage the workers to express their emotions in any possible way. As emojis provide a convenient way to express emotions in textual communications, online work platforms may take measures to enable and promote the use of emojis. Second, negative emotions (e.g., deterrence such as anxiety and fear, withdrawal such as sadness and shame, and antagonism such as anger and hate) are particularly useful signals for negative work experiences and outcomes (e.g., risks of burnouts and dropouts).  
Employers should pay particular attention to related emojis or similar signals of remote workers. Third, there are multiple paths of using emojis, and our results show that the number of emoji posts, the number of emojis, and the variety of emojis are good indicators of (low) dropout risk with the emotional scores of emojis controlled. Thus users should be encouraged to use a larger variety of emojis and use emojis for multiple purposes rather than just expressing emotions. Fourth, obsession is another signal to watch from the use of emojis or other activities -- obsessive passion leads to burnout, and burnout leads to dropout. On the opposite, using a variety of emojis may reduce the stress of work and distract workers from obsessive passion and potential burnouts. Platforms, worker communities, and organizations with remote workers may use various means to encourage workers to adopt a wide range of emojis, e.g., through building an emoji recommender system. 

Finally, our results echo many theories and findings in worker psychology and organization science, which are typically done using small scale surveys or experiments. The use of emojis introduces noisier but much larger scale observations of emotions at work, which provides a promising new instrument for such studies.  

Note that the main goal of our work is not to optimize the accuracy of dropout prediction. Instead, we utilize the task to verify the predictive power of emotions expressed by remote workers. Rather than accuracy, we care more about the explainability of the results, especially the role of emojis. A study of a similar goal is \cite{weisberg1994measuring}, which examines the efficacy of physical exhaustion, mental exhaustion, and emotional exhaustion (measured by survey) in predicting the intention of school teachers to leave their jobs due to burnout. With an expanded set of features describing the profile and activities of individuals and advanced machine learning models, such a prediction task could achieve a much high accuracy (e.g., ~\cite{xing2019dropout}). 

\subsection*{Limitations}
We notice some limitations in our analysis. First, although confounding factors like the activity levels and programming languages are controlled for in the regressions and predictions, there are more factors that we are not able to observe in our dataset, such as demographics (e.g., gender, age, and country) of the developers. It requires further analysis to determine whether the conclusions we draw are causal. Second, the measurements of working status are limited by the platform and the dataset. Since the use of GitHub only partially reflects the tasks of a developer, we are unable to observe their activities offline or in private repositories. The prediction outcome only refers to dropouts from public repositories of GitHub but does not necessarily reflect dropouts from their job. Additionally, our approach to identifying emotions of emojis is primitive, as it works in a collective sense without considering the context of individual use of emojis. Further work may drill down and distinguish different purposes of emojis in individual contexts and compare with our conclusion. Finally, our analysis benefits from the established norm of using online collaborative platform by developers. Readers should generalize our conclusions to other domains with caution. 

\section*{Conclusion}\label{sec:conclusion}
We propose to use emojis in online collaborative platforms as sensors for tracking the emotions of remote workers. In the context of GitHub, we discover patterns of how developers use emojis in work-related activities and communications. We find significant coefficients of working status variables in regressions of emoji usage in the same year. We show that with emoji usage as features alone, standard machine learning models can predict future dropouts of developers at a satisfactory accuracy (above 70\% overall and up to 75\% in classification accuracy; above 75\% overall and up to 82\% in AUC). Emotion-related features appear to be highly important, while relations between other purposes of emoji usage and dropout could also exist. Through the prediction analysis, we are able to provide practical implications for using emojis to sense and reduce the risk of negative work experiences and outcomes. Future work should distinguish other purposes of emoji use in the prediction tasks and investigate the causal effect of using emojis to improve work outcomes. It will also be intriguing to test these findings in other remote work platforms and other remote communication contexts, such as online education. 






\section*{Supporting Information Captions}


\paragraph*{{S1 Appendix}}
\label{S1_appendix}

\section*{Acknowledgments}

The authors would like to thank the reviewers of this manuscript for their constructive comments and Dr. Xuanzhe Liu from Peking University for valuable suggestions.  Qiaozhu Mei and Xuan Lu's work is in part supported by the National Science Foundation under grant number 1633370.

\nolinenumbers

%
%
%


\section*{Appendix}
\paragraph*{1. Programming language selection}
\label{S1_lang}
We adopt GitHub's definition of ``a language.'' We calculate the popularity of a language based on the number of repositories that identify it as their main languages based on pull request events, pull request, and review comment events in the GHArchive dataset. These languages are (in decreasing popularity): JavaScript, Python, Java, C++, GO, HTML, PHP, Ruby, TypeScript, C\#, C, CSS, Shell, Rust, Scala, Swift, Objective-C, PowerShell, Jupyter Notebook, and Kotlin.

\paragraph*{2. Regression configurations}
The independent variables in the first regression (1) are all from category D1 (activity level); the independent variables in the second regression (2) are from D1 and D2, and so on. Finally, in the last model (6), we include all categories of variables (D1-5) as well as the control variables (platform age and programming languages).

\paragraph*{3. Regression results}
\label{S3_regression}
We report the full results of the regressions in Table~\ref{tab:regression_general_log}, Table~\ref{tab:regression_general_prop}, and Table~\ref{tab:regression_ee}.

\begin{table}[!ht]
\tiny
\begin{adjustwidth}{-2.25in}{0in}
\centering
\caption{{\bf OLS Regressions of Emoji Usage (\#Emoji Posts) on Working Status.}}
\begin{tabular}{lccccccllc}
\toprule
\multicolumn{10}{c}{Dependent Variable: Number of Emoji Posts (Log Scale)}\\\midrule
 & (1) & (2) & (3) & (4) & (5) & (6) & & & (6) (cont.)\\
\midrule
const	&	3.3939***	&	3.3497***	&	3.0106***	&	2.6682***	&	2.4833***	&	2.0552***	&	&	platform age (log)	&	0.0373***	\\
	&	(0.0988)	&	(0.0989)	&	(0.0950)	&	(0.0925)	&	(0.0905)	&	(0.0899)	&	&		&	(0.0011)	\\
\textit{avg.} working hours&	0.0322***	&	0.0335***	&	0.0355***	&	0.0468***	&	0.0240***	&	0.0334***	&	&	C	&	-0.1719***	\\
	&	(0.0018)	&	(0.0018)	&	(0.0019)	&	(0.0019)	&	(0.0018)	&	(0.0018)	&	&		&	(0.0063)	\\
\textit{avg.} length of working sessions (in hour)&	0.0002	&	0.0002	&	0.0006***	&	0.0004***	&	0.0003**	&	0.0002*	&	&	CSS	&	0.0986***	\\
	&	(0.0002)	&	(0.0002)	&	(0.0001)	&	(0.0001)	&	(0.0001)	&	(0.0001)	&	&		&	(0.0069)	\\
length of off segment (threshold = 4) &	-0.0366***	&	-0.0365***	&	-0.0455***	&	-0.0393***	&	-0.0265***	&	-0.0254***	&	&	C\#	&	-0.0230*** 	\\
	&	(0.0002)	&	(0.0002)	&	(0.0002)	&	(0.0002)	&	(0.0002)	&	(0.0002)	&	&		&	(0.0053)	\\
length of off segment (threshold = 16) &	-0.0391***	&	-0.0390***	&	-0.0308***	&	-0.0291***	&	-0.0324***	&	-0.0320***	&	&	C++	&	-0.1550***	\\
	&	(0.0004)	&	(0.0004)	&	(0.0004)	&	(0.0004)	&	(0.0004)	&	(0.0004)	&	&		&	(0.0050)	\\
length of off segment (threshold = 32) &	-0.0350***	&	-0.0349***	&	-0.0269***	&	-0.0246***	&	-0.0229***	&	-0.0231***	&	&	GCC	&	0.5722***	\\
	&	(0.0007)	&	(0.0007)	&	(0.0006)	&	(0.0006)	&	(0.0006)	&	(0.0006)	&	&		&	(-0.0122)	\\
length of off segment (threshold = 64) &	-0.0482***	&	-0.0477***	&	-0.0351***	&	-0.0326***	&	-0.0302***	&	-0.0293***	&	&	Go	&	-0.0224***	\\
	&	(0.0011)	&	(0.0011)	&	(0.0010)	&	(0.0010)	&	(0.0010)	&	(0.0010)	&	&		&	(-0.0063)	\\
length of off segment (threshold = 128) &	-0.0055**	&	-0.0051**	&	-0.0105***	&	-0.0115***	&	-0.0168***	&	-0.0153***	&	&	HTML	&	0.0550***	\\
	&	(0.0024)	&	(0.0024)	&	(0.0023)	&	(0.0023)	&	(0.0022)	&	(0.0022)	&	&		&	(-0.0048)	\\
length of off segment (threshold = 256) &	0.0322***	&	0.0329***	&	0.0233***	&	0.0255***	&	0.0104**	&	0.0145***	&	&	Java	&	-0.1088***	\\
	&	(0.0046)	&	(0.0046)	&	(0.0044)	&	(0.0042)	&	(0.0041)	&	(0.0041)	&	&		&	(-0.0039)	\\
monthly trend of \#working days&		&	-0.0206***	&	-0.0166***	&	-0.0103***	&	-0.0101***	&	-0.0069***	&	&	JavaScript	&	0.1134***	\\
	&		&	(0.0017)	&	(0.0016)	&	(0.0016)	&	(0.0015)	&	(0.0015)	&	&		&	(-0.0029)	\\
monthly trend of \textit{avg.} working hours&		&	0.0130***	&	0.0132***	&	0.0129***	&	0.0098***	&	0.0117***	&	&	Jupyter Notebook	&	-0.0953***	\\
	&		&	(0.0016)	&	(0.0015)	&	(0.0015)	&	(0.0014)	&	(0.0014)	&	&		&	(-0.0132)	\\
monthly trend of \textit{avg.} length of working sessions&		&	0.0002	&	-0.0019	&	-0.0012	&	-0.0001	&	0.0019	&	&	Objective C	&	0.0169	\\
	&		&	(0.0015)	&	(0.0014)	&	(0.0014)	&	(0.0013)	&	(0.0013)	&	&		&	(-0.0111)	\\
monthly trend of \#working sessions with >1 event&		&	-0.0068***	&	-0.0101***	&	-0.0099***	&	-0.0077***	&	-0.0068***	&	&	PHP	&	0.0170***	\\
	&		&	(0.0017)	&	(0.0016)	&	(0.0016)	&	(0.0015)	&	(0.0015)	&	&		&	(-0.0050)	\\
monthly trend of \#posts &		&	0.0059***	&	0.0128***	&	0.0085***	&	0.0105***	&	0.0113***	&	&	PowerShell	&	0.4798***	\\
	&		&	(0.0013)	&	(0.0013)	&	(0.0012)	&	(0.0012)	&	(0.0012)	&	&		&	(-0.0078)	\\
\textit{prop.} pull request events&		&		&	0.7475***	&	0.8098***	&	0.2979***	&	0.2079***	&	&	Python	&	-0.0658***	\\
	&		&		&	(0.0164)	&	(0.0167)	&	(0.0166)	&	(0.0166)	&	&		&	(-0.0037)	\\
\textit{prop.} push events&		&		&	-0.2224***	&	-0.2253***	&	-0.3261***	&	-0.3165***	&	&	Ruby	&	0.1613***	\\
	&		&		&	(0.0048)	&	(0.0047)	&	(0.0047)	&	(0.0047)	&	&		&	(-0.0060)	\\
\textit{prop.} working sessions with pull request events &		&		&	0.0288***	&	0.0718***	&	0.4902***	&	0.5022***	&	&	Rust	&	0.1202***	\\
	&		&		&	(0.0074)	&	(0.0074)	&	(0.0077)	&	(0.0077)	&	&		&	(-0.0136)	\\
\textit{prop.} working days with comment events &		&		&	0.5437***	&	0.2777***	&	0.3839***	&	0.4069***	&	&	Scala	&	-0.0681***	\\
	&		&		&	(0.0034)	&	(0.0045)	&	(0.0045)	&	(0.0045)	&	&		&	(-0.0139)	\\
\textit{prop.} working days with issue events&		&		&	0.1998***	&	0.2637***	&	0.4348***	&	0.3983***	&	&	Shell	&	-0.0205***	\\
	&		&		&	(0.0038)	&	(0.0049)	&	(0.0049)	&	(0.0049)	&	&		&	(-0.0079)	\\
\textit{prop.} pull requests&		&		&		&	-0.1556***	&	-0.1428***	&	-0.1269***	&	&	Swift	&	0.2226***	\\
	&		&		&		&	(0.0051)	&	(0.0050)	&	(0.0049)	&	&		&	(-0.0098)	\\
\textit{prop.} issue comments&		&		&		&	0.0902***	&	0.0223***	&	0.0176***	&	&	TypeScript	&	0.1006***	\\
	&		&		&		&	(0.0053)	&	(0.0052)	&	(0.0051)	&	&		&	(-0.0059)	\\
\textit{prop.} pull request comments&		&		&		&	0.6779***	&	0.5859***	&	0.5622***	&	&		&		\\
	&		&		&		&	(0.0071)	&	(0.0070)	&	(0.0069)	&	&		&		\\
\textit{prop.} pull request review comments&		&		&		&	0.7706***	&	0.6380***	&	0.6093***	&	&		&		\\
	&		&		&		&	(0.0079)	&	(0.0078)	&	(0.0078)	&	&		&		\\
\textit{prop.} commit comments&		&		&		&	-0.0087	&	0.0355***	&	0.0559***	&	&		&		\\
	&		&		&		&	(0.0083)	&	(0.0081)	&	(0.0080)	&	&		&		\\
entropy of weekdays&		&		&		&		&	0.3188***	&	0.3030***	&	&		&		\\
	&		&		&		&		&	(0.0021)	&	(0.0021)	&	&		&		\\
variables about backgrounds	&		&		&		&		&		&	see right	&	&		&		\\
\midrule																		
\# Obs.	&	529616	&	529616	&	529616	&	529616	&	529616	&	529616	&	&		&		\\
R-squared	&	0.3548	&	0.3552	&	0.4140	&	0.4466	&	0.4704	&	0.4845	&	&		&		\\
Adj. R-squared	&	0.3549	&	0.3552	&	0.4141	&	0.4466	&	0.4704	&	0.4846	&	&		&		\\ 
\bottomrule
\label{tab:regression_general_log}
\end{tabular}
\begin{flushleft} \textit{Notes}: Standard errors in parentheses.\\
Significant at the: *** 1\%, ** 5\%, or * 10\% level.
\end{flushleft}
\end{adjustwidth}
\end{table}

\begin{table}[!ht]
\tiny
\begin{adjustwidth}{-2.25in}{0in}
\centering
\caption{{\bf OLS Regressions of Emoji Usage (\textit{prop.} Emoji Posts) on Working Status.}}
\begin{tabular}{lccccccllc}
\toprule
\multicolumn{10}{c}{Dependent Variable: Proportion of Emoji Posts}\\\midrule
 & (1) & (2) & (3) & (4) & (5) & (6) & & & (6) (cont.)\\
\midrule
const		&	0.5533***	&	0.5617***	&	0.3244***	&	0.2649***	&	0.2867***	&	0.1903***	&	&	platform age (log)	&	0.0032***	\\
		&	(0.0313)	&	(0.0313)	&	(0.0313)	&	(0.0311)	&	(0.0310)	&	(0.0301)	&	&		&	(0.0004)	\\
\textit{avg.} working hours	&	-0.0223***	&	-0.0227***	&	-0.0068***	&	-0.0037***	&	-0.0010	&	0.0024***	&	&	C	&	-0.0348***	\\
		&	(0.0006)	&	(0.0006)	&	(0.0006)	&	(0.0006)	&	(0.0006)	&	(0.0006)	&	&		&	(0.0021)	\\
\textit{avg.} length of working sessions (in hour)	&	0.0003***	&	0.0003***	&	0.0003***	&	0.0002***	&	0.0002***	&	0.0002***	&	&	CSS	&	0.0227***	\\
		&	(0.0000)	&	(0.0000)	&	(0.0000)	&	(0.0000)	&	(0.0000)	&	(0.0000)	&	&		&	(0.0023)	\\
length of off segment (threshold = 4)	&	0.0026***	&	0.0026***	&	0.0023***	&	0.0030***	&	0.0015***	&	0.0016***	&	&	C\#	&	-0.0032*	\\
		&	(0.0001)	&	(0.0001)	&	(0.0001)	&	(0.0001)	&	(0.0001)	&	(0.0001)	&	&		&	(0.0018)	\\
length of off segment (threshold = 16)	&	0.0001	&	0.0001	&	0.0001	&	0.0003**	&	0.0007***	&	0.0008***	&	&	C++	&	-0.0324***	\\
		&	(0.0001)	&	(0.0001)	&	(0.0001)	&	(0.0001)	&	(0.0001)	&	(0.0001)	&	&		&	(0.0017)	\\
length of off segment (threshold = 32)	&	-0.0005**	&	-0.0006***	&	-0.0006***	&	-0.0003*	&	-0.0005***	&	-0.0006***	&	&	GCC	&	0.4173***	\\
		&	(0.0002)	&	(0.0002)	&	(0.0002)	&	(0.0002)	&	(0.0002)	&	(0.0002)	&	&		&	(0.0041)	\\
length of off segment (threshold = 64)	&	-0.0026***	&	-0.0027***	&	-0.0020***	&	-0.0014***	&	-0.0017***	&	-0.0014***	&	&	Go	&	-0.0087***	\\
		&	(0.0003)	&	(0.0003)	&	(0.0003)	&	(0.0003)	&	(0.0003)	&	(0.0003)	&	&		&	(0.0021)	\\
length of off segment (threshold = 128)	&	-0.0087***	&	-0.0088***	&	-0.0047***	&	-0.0037***	&	-0.0031***	&	-0.0022***	&	&	HTML	&	0.0119***	\\
		&	(0.0008)	&	(0.0008)	&	(0.0008)	&	(0.0008)	&	(0.0008)	&	(0.0007)	&	&		&	(0.0016)	\\
length of off segment (threshold = 256)	&	-0.0074***	&	-0.0074***	&	-0.0014	&	-0.0003	&	0.0015	&	0.0029**	&	&	Java	&	-0.0328***	\\
		&	(0.0014)	&	(0.0014)	&	(0.0014)	&	(0.0014)	&	(0.0014)	&	(0.0014)	&	&		&	(0.0013)	\\
monthly trend of \#working days	&		&	0.0027***	&	0.0022***	&	0.0028***	&	0.0028***	&	0.0014***	&	&	JavaScript	&	0.0140***	\\
		&		&	(0.0005)	&	(0.0005)	&	(0.0005)	&	(0.0005)	&	(0.0005)	&	&		&	(0.0010)	\\
monthly trend of \textit{avg.} working hours	&		&	0.0022***	&	0.0022***	&	0.0022***	&	0.0025***	&	0.0014***	&	&	Jupyter Notebook	&	-0.0308***	\\
		&		&	(0.0005)	&	(0.0005)	&	(0.0005)	&	(0.0005)	&	(0.0005)	&	&		&	(0.0044)	\\
monthly trend of \textit{avg.} length of working sessions	&		&	-0.0023***	&	-0.0022***	&	-0.0023***	&	-0.0024***	&	-0.0013***	&	&	Objective C	&	-0.0043	\\
		&		&	(0.0005)	&	(0.0005)	&	(0.0005)	&	(0.0005)	&	(0.0004)	&	&		&	(0.0037)	\\
monthly trend of \#working sessions with >1 event	&		&	-0.0061***	&	-0.0055***	&	-0.0053***	&	-0.0056***	&	-0.0037***	&	&	PHP	&	-0.0151***	\\
		&		&	(0.0005)	&	(0.0005)	&	(0.0005)	&	(0.0005)	&	(0.0005)	&	&		&	(0.0017)	\\
monthly trend of \#posts	&		&	0.0087***	&	0.0081***	&	0.0078***	&	0.0075***	&	0.0064***	&	&	PowerShell	&	0.4256***	\\
		&		&	(0.0004)	&	(0.0004)	&	(0.0004)	&	(0.0004)	&	(0.0004)	&	&		&	(0.0026)	\\
\textit{prop.} pull request events	&		&		&	-0.1669***	&	-0.1606***	&	-0.1003***	&	-0.1090***	&	&	Python	&	-0.0231***	\\
		&		&		&	(0.0054)	&	(0.0056)	&	(0.0057)	&	(0.0056)	&	&		&	(0.0012)	\\
\textit{prop.} push events	&		&		&	-0.0569***	&	-0.0617***	&	-0.0498***	&	-0.0496***	&	&	Ruby	&	0.0268***	\\
		&		&		&	(0.0016)	&	(0.0016)	&	(0.0016)	&	(0.0016)	&	&		&	(0.0020)	\\
\textit{prop.} working sessions with pull request events 	&		&		&	-0.0072***	&	0.0055**	&	-0.0438***	&	-0.0476***	&	&	Rust	&	-0.0003	\\
		&		&		&	(0.0025)	&	(0.0025)	&	(0.0026)	&	(0.0026)	&	&		&	(0.0046)	\\
\textit{prop.} working days with comment events	&		&		&	-0.0679***	&	-0.0843***	&	-0.0968***	&	-0.0871***	&	&	Scala	&	-0.0283***	\\
		&		&		&	(0.0011)	&	(0.0015)	&	(0.0015)	&	(0.0015)	&	&		&	(0.0047)	\\
\textit{prop.} working days with issue events	&		&		&	0.0267***	&	0.0052***	&	-0.0150***	&	-0.0487***	&	&	Shell	&	-0.0082***	\\
		&		&		&	(0.0013)	&	(0.0016)	&	(0.0017)	&	(0.0016)	&	&		&	(0.0026)	\\
\textit{prop.} pull requests	&		&		&		&	-0.0637***	&	-0.0652***	&	-0.0606***	&	&	Swift	&	0.0466***	\\
		&		&		&		&	(0.0017)	&	(0.0017)	&	(0.0017)	&	&		&	(0.0033)	\\
\textit{prop.} issue comments	&		&		&		&	-0.0479***	&	-0.0399***	&	-0.0369***	&	&	TypeScript	&	0.0083***	\\
		&		&		&		&	(0.0018)	&	(0.0018)	&	(0.0017)	&	&		&	(0.0020)	\\
\textit{prop.} pull request comments	&		&		&		&	0.1339***	&	0.1447***	&	0.1384***	&	&		&		\\
		&		&		&		&	(0.0024)	&	(0.0024)	&	(0.0023)	&	&		&		\\
\textit{prop.} pull request review comments	&		&		&		&	-0.0363***	&	-0.0206***	&	-0.0196***	&	&		&		\\
		&		&		&		&	(0.0027)	&	(0.0027)	&	(0.0026)	&	&		&		\\
\textit{prop.} commit comments	&		&		&		&	-0.0258***	&	-0.0310***	&	-0.0284***	&	&		&		\\
		&		&		&		&	(0.0028)	&	(0.0028)	&	(0.0027)	&	&		&		\\
entropy of weekdays	&		&		&		&		&	-0.0376***	&	-0.0337***	&	&		&		\\
		&		&		&		&		&	(0.0007)	&	(0.0007)	&	&		&		\\
variables about backgrounds	&		&		&		&		&		&	see right	&	&		&		\\
\midrule																			
\# Obs.		&	529616	&	529616	&	529616	&	529616	&	529616	&	529616	&	&		&		\\
R-squared		&	0.0161	&	0.0176	&	0.0340	&	0.0538	&	0.0588	&	0.1255	&	&		&		\\
Adj. R-squared		&	0.0161	&	0.0176	&	0.0340	&	0.0538	&	0.0588	&	0.1255	&	&		&		\\ 
\bottomrule
\label{tab:regression_general_prop}
\end{tabular}
\begin{flushleft} \textit{Notes}: Standard errors in parentheses.\\
Significant at the: *** 1\%, ** 5\%, or * 10\% level.
\end{flushleft}
\end{adjustwidth}
\end{table}

\begin{table}[!ht]
\tiny
\begin{adjustwidth}{-2.25in}{0in}
\centering
\caption{{\bf OLS Regressions of Emoji Usage (\textit{prop.} Emotional Emojis) on Working Status.}}
\begin{tabular}{lccccccllc}
\toprule
\multicolumn{10}{c}{Dependent Variable: Prop. emotional emojis}\\\midrule
 & (1) & (2) & (3) & (4) & (5) & (6) & & & (6) (cont.)\\
 \midrule

const                      & 1.4479***                & 1.4788***                 & 1.2302***                  & 1.1711***                   & 1.2646***                    & 0.9679***    && platform age (log)          & 0.0030** \\
                           & (0.0693)                 & (0.0693)                  & (0.0697)                   & (0.0699)                    & (0.0706)                     & (0.0710)     &&                             & (0.0012)  \\
\textit{avg.} working hours & -0.0586***               & -0.0599***                & -0.0444***                 & -0.0384***                  & -0.0403***                   & -0.0294***   && C                           & -0.0103*               \\
                           & (0.0017)                 & (0.0017)                  & (0.0018)                   & (0.0018)                    & (0.0018)                     & (0.0018)     &&                             & (0.0062)               \\
\textit{avg.} length of working sessions & 0.0002**                 & 0.0002**                  & 0.0002                     & 0.0001                      & 0.0001                       & 0.0001       && CSS                         & -0.0014 \\
                           & (0.0001)                 & (0.0001)                  & (0.0001)                   & (0.0001)                    & (0.0001)                     & (0.0001)     &&                             & (0.0061) \\
length of off segment (threshold = 1) & 0.0029***                & 0.0028***                 & 0.0017***                  & 0.0019***                   & 0.0002                       & -0.0006*     && C\#                    & 0.1203*** \\
                           & (0.0002)                 & (0.0002)                  & (0.0002)                   & (0.0002)                    & (0.0003)                     & (0.0003)     &&                             & (0.0051)   \\
length of off segment (threshold = 4) & -0.0019***               & -0.0019***                & -0.0019***                 & -0.0021***                  & -0.0023***                   & -0.0016***   && C++                         & -0.0209*** \\
                           & (0.0002)                 & (0.0002)                  & (0.0002)                   & (0.0002)                    & (0.0002)                     & (0.0002)     &&                             & (0.0048)\\
length of off segment (threshold = 16) & -0.0021***               & -0.0021***                & -0.0018***                 & -0.0018***                  & -0.0017***                   & -0.0013***   && GCC                         & 0.6059*** \\
                           & (0.0003)                 & (0.0003)                  & (0.0003)                   & (0.0003)                    & (0.0003)                     & (0.0003)     &&                             & (0.0088) \\
length of off segment (threshold = 32) & -0.0002                  & -0.0004                   & -0.0001                    & -0.0002                     & -0.0002                      & -0.0000      && Go                          & -0.0027 \\
                           & (0.0004)                 & (0.0004)                  & (0.0004)                   & (0.0004)                    & (0.0004)                     & (0.0004)     &&                             & (0.0052)  \\
length of off segment (threshold = 64) & -0.0059***               & -0.0062***                & -0.0049***                 & -0.0048***                  & -0.0050***                   & -0.0041***   && HTML                        & 0.0196***  \\
                           & (0.0007)                 & (0.0007)                  & (0.0007)                   & (0.0007)                    & (0.0007)                     & (0.0007)     &&                             & (0.0046) \\
length of off segment (threshold = 128) & -0.0113***               & -0.0115***                & -0.0075***                 & -0.0062***                  & -0.0066***                   & -0.0043***   && Java                        & -0.0090**  \\
                           & (0.0016)                 & (0.0016)                  & (0.0016)                   & (0.0016)                    & (0.0016)                     & (0.0016)     &&                             & (0.0040) \\
length of off segment (threshold = 256) & -0.0210***               & -0.0214***                & -0.0155***                 & -0.0128***                  & -0.0133***                   & -0.0091***   && JavaScript                  & 0.0027 \\
                           & (0.0030)                 & (0.0030)                  & (0.0030)                   & (0.0030)                    & (0.0030)                     & (0.0030)     &&                             & (0.0026)\\
monthly trend of \#events &                          & 0.0058***                 & 0.0062***                  & 0.0066***                   & 0.0065***                    & 0.0042**     && Jupyter Notebook           & 0.0076  \\
                           &                          & (0.0018)                  & (0.0018)                   & (0.0018)                    & (0.0018)                     & (0.0018)     &&                             & (0.0136) \\
monthly trend of \#working days &                          & 0.0098***                 & 0.0081***                  & 0.0076***                   & 0.0076***                    & 0.0059***    && Objective C                & -0.0031\\
                           &                          & (0.0015)                  & (0.0015)                   & (0.0015)                    & (0.0015)                     & (0.0015)     &&                             & (0.0103)\\
monthly trend of \textit{avg.} working hours &                          & 0.0061***                 & 0.0054***                  & 0.0055***                   & 0.0055***                    & 0.0042***    && PHP                         & -0.0205*** \\
                           &                          & (0.0013)                  & (0.0013)                   & (0.0013)                    & (0.0013)                     & (0.0013)     &&                             & (0.0046) \\
monthly trend of \textit{avg.} length of working sessions &                          & -0.0045***                & -0.0039***                 & -0.0039***                  & -0.0039***                   & -0.0025**    && PowerShell                  & 0.3107***\\
                           &                          & (0.0012)                  & (0.0012)                   & (0.0012)                    & (0.0012)                     & (0.0011)     &&                             & (0.0060) \\
monthly trend of \#working sessions with >1 event  &                          & -0.0170***                & -0.0151***                 & -0.0150***                  & -0.0149***                   & -0.0117***   && Python                      & -0.0020 \\
                           &                          & (0.0016)                  & (0.0016)                   & (0.0016)                    & (0.0016)                     & (0.0015)     &&                             & (0.0035) \\
monthly trend of \#posts &                          & 0.0139***                 & 0.0125***                  & 0.0126***                   & 0.0126***                    & 0.0113***    && Ruby                        & -0.0063 \\
                           &                          & (0.0012)                  & (0.0012)                   & (0.0012)                    & (0.0012)                     & (0.0012)     &&                             & (0.0049)\\
\textit{prop.} comment events  &                          &                           & -0.0968***                 & -0.0680***                  & -0.0696***                   & -0.0491***   && Rust                        & 0.0432*** \\
                           &                          &                           & (0.0044)                   & (0.0058)                    & (0.0058)                     & (0.0059)     &&                             & (0.0102)\\
\textit{prop.} working days with issue events &                          &                           & 0.0595***                  & 0.0038                      & -0.0061                      & -0.0781***   && Scala                       & -0.0157 \\
                           &                          &                           & (0.0042)                   & (0.0054)                    & (0.0055)                     & (0.0056)     &&                            & (0.0116)\\
\textit{prop.} pull request events &                          &                           & -0.3498***                 & -0.2256***                  & -0.2212***                   & -0.2143***   && Shell                       & -0.0024 \\
                           &                          &                           & (0.0103)                   & (0.0137)                    & (0.0137)                     & (0.0137)     &&                             & (0.0070)  \\
\textit{prop.} push events &                          &                           & -0.0532***                 & -0.0443***                  & -0.0402***                   & -0.0307***   && Swift                       & 0.0463*** \\
                           &                          &                           & (0.0050)                   & (0.0051)                    & (0.0051)                     & (0.0050)     &&                             & (0.0082)\\
\textit{prop.} commit comments &                          &                           &                            & -0.1130***                  & -0.1160***                   & -0.0979***   && TypeScript                  & 0.0142***  \\
                           &                          &                           &                            & (0.0110)                    & (0.0110)                     & (0.0109)     &&                             & (0.0051) \\
\textit{prop.} issue comments &                          &                           &                            & -0.0459***                  & -0.0430***                   & -0.0228***   &&&                  \\
                           &                          &                           &                            & (0.0060)                    & (0.0060)                     & (0.0059)     &&&                  \\
\textit{prop.} pull request comments &                          &                           &                            & -0.0890***                  & -0.0900***                   & -0.0857***   &&&                  \\
                           &                          &                           &                            & (0.0070)                    & (0.0070)                     & (0.0069)     && &                 \\
\textit{prop.} pull requests&                          &                           &                            & -0.0811***                  & -0.0840***                   & -0.0739***   && &                 \\
                           &                          &                           &                            & (0.0065)                    & (0.0065)                     & (0.0064)     && &                 \\
\textit{prop.} pull request review comments &                          &                           &                            & -0.1117***                  & -0.1091***                   & -0.1056***   && &                 \\
                           &                          &                           &                            & (0.0076)                    & (0.0076)                     & (0.0076)     && &                 \\
entropy of weekdays&                          &                           &                            &                             & -0.0298***                   & -0.0034      && &                 \\
                           &                          &                           &                            &                             & (0.0033)                     & (0.0033)     && &                 \\
variables about backgrounds   &     &     &     &     &     &  see right   &  &     &     \\
\midrule
\# Obs.     &  264808   &  264808   &  264808   &  264808   &  264808   &  264808   &  &     &     \\
R-squared                  & 0.0064                   & 0.0084                    & 0.0159                     & 0.0173                      & 0.0176                       & 0.0443       && &                 \\
                           & 0.0064                   & 0.0084                    & 0.0160                     & 0.0174                      & 0.0177                       & 0.0444       && &                 \\
\bottomrule
\label{tab:regression_ee}
\end{tabular}
\begin{flushleft} \textit{Notes}: Standard errors in parentheses.\\
Significant at the: *** 1\%, ** 5\%, or * 10\% level.
\end{flushleft}
\end{adjustwidth}
\end{table}

\begin{figure}[!h]
\includegraphics[width=0.5\linewidth]{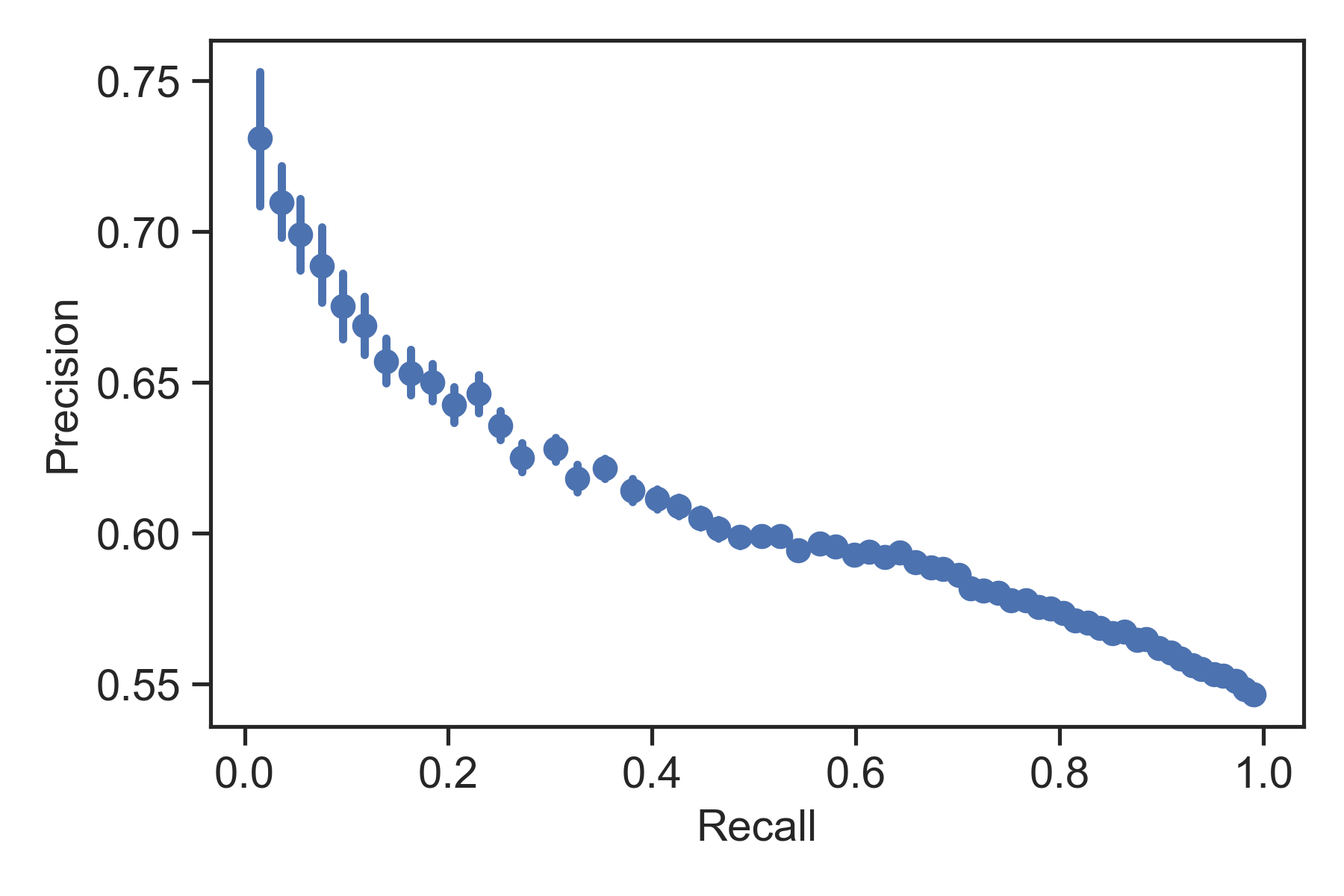}
\caption{{\bf Precision-recall curve for top 5\% active users (by \#working days).} The values of recall and precision at different threshold levels are binned for the 30 sampled datasets.
}
\label{fig:pred_precision_recall}
\end{figure}

\paragraph*{4. Precision-recall curve}
\label{S4_curve}
Fig~\ref{fig:pred_precision_recall} presents the precision-recall curve of GBDT for the 30 datasets with the $k=5$ activity level (by number of working days). This curve demonstrates the precision values for units ranked by their predicted probability of dropping out.

\begin{figure}[!h]
\includegraphics[width=1\linewidth]{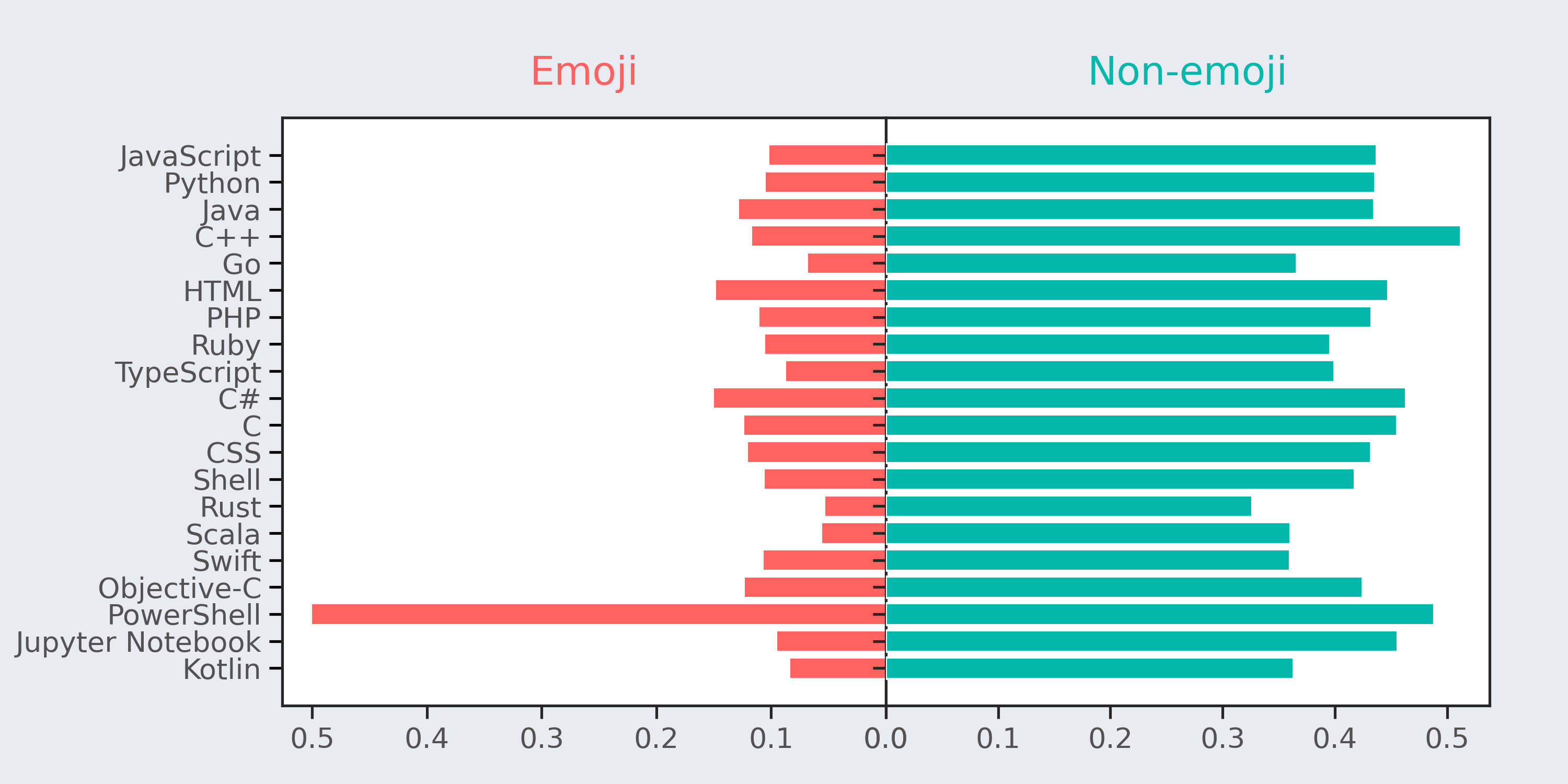}
\caption{{\bf Dropout ratio for developers with different primary programming languages.} Emoji: developers who used at least one emoji in 2018. Non-emoji: developers who did not use any emoji in 2018.
}
\label{fig:lang_dropout}
\end{figure}

\paragraph*{5. Dropout ratio vs. programming languages}
\label{S5_ratio}
To demonstrate the relation between the primary programming language and the working status outcome, we plot the distribution of dropout rates in the 20 most popular programming languages in Fig~\ref{fig:lang_dropout}. Variance could be observed among programming languages from both the developers who used emojis and the others in 2018. 
\end{document}